# Human-Enhanced Loop Modeling (HELM): Agent-Based Finite Element Modeling of Concrete Bridge Barriers

Quankai Wang[1,*]; Yulin Xie[2,*]; Tongfei Yang[3]; Minghui Cheng, A.M.ASCE[4]; and Ran Cao, M.ASCE[5]

**Abstract:** Finite element (FE) modeling of safety-critical infrastructure such as bridge barriers requires high-fidelity nonlinear dynamic analysis, yet the current FE modeling process remains labor-intensive and lacks automation. This paper presents the Human-Enhanced Loop Modeling (HELM) framework, a collaborative human-agent protocol that decomposes long-sequence finite element modeling into discrete, visually verifiable checkpoints across geometry generation, boundary condition definition, and material assignment. The framework is demonstrated through a 20-case matrix of reinforced concrete bridge barriers under MASH TL-4 and TL-5 lateral loading conditions, interfacing specialized agents with two widely used commercial FE software platforms, i.e., ANSYS and LS-PrePost. Experimental results show that HELM improves the baseline autonomous modeling success rate from 20% to 75%, with agent-level pass rates for geometry and boundary condition tasks approximately doubling. Error analysis reveals that spatial reasoning and algebraic logic limitations constitute the primary failure modes, underscoring the value of structured human-in-the-loop intervention for modeling automation. The complete agent design code and prompts are open-sourced and can be accessed at: https://github.com/SimAgentDev/Ansys-LSPP-AgentKit.

**Author keywords:** Large language models (LLM); LLM agents; Human-in-the-loop (HITL); Finite element analysis (FEA); Bridge barriers.

[1] Research Assistant, College of Civil Engineering, Hunan University, Changsha, 410082, China. E-mail: wangquankai@hnu.edu.cn
[2] Research Assistant, College of Civil Engineering, Hunan University, Changsha, 410082, China. E-mail: ylxie2024@hnu.edu.cn
[3] Research Assistant, College of Civil Engineering, Hunan University, Changsha, 410082, China. E-mail: yangtf@hnu.edu.cn
[4] Assistant Professor, Department of Civil and Architectural Engineering, University of Miami, Coral Gables, FL 33146; School of Architecture, University of Miami, Coral Gables, FL 33146. E-mail: minghui.cheng@miami.edu
[5] Associate Professor, College of Civil Engineering, Hunan University, Changsha 410082, China (corresponding author). Email: rcao@hnu.edu.cn
*Equal contribution.

## Introduction

Bridge barriers are critical roadside safety components intended to prevent vehicles from leaving the roadway or crossing into opposing traffic. Their design is governed by evolving standards from the American Association of State Highway and Transportation Officials (AASHTO-LRFD 2024), particularly the transition from NCHRP Report 350 (Ross et al. 1993) to the Manual for Assessing Safety Hardware (MASH 2016). This transition reflects increased impact severity requirements, including higher vehicle weights, speeds, and impact angles. For example, Test Level 4 (TL-4) conditions have seen a substantial increase in design loads, significantly raising the demands placed on structural performance.

Accurate evaluation of bridge barrier performance under these conditions requires nonlinear dynamic finite element analysis (FEA), typically implemented in specialized solvers such as ANSYS (Ansys 2024) and LS-DYNA (Hallquist 2006). It should be noted that the modeling process of the barrier system is inherently complex, involving detailed representations of reinforced concrete components, deck overhangs, and reinforcement layouts. Meanwhile, simplified design approaches, such as AASHTO yield line methods (AASHTO-LRFD 2024), have been criticized for their inability to capture realistic cracking patterns and failure mechanisms observed in experiments (Cao et al. 2020, 2021; Frosch and Morel 2016; Steelman et al. 2023). To address this issue, high-fidelity simulations offer a path toward performance-based design, while its adoption is hindered by the substantial effort required to construct detailed models, particularly in defining reinforcement geometry and coordinating across multiple software platforms.

In parallel with these domain-specific challenges, recent advancements in large language models (LLMs) and AI agents present a transformative opportunity for engineering workflows. LLMs such as GPT-4o (OpenAI 2024), Llama-3.3 (Grattafiori 2024), and DeepSeek-V3 (DeepSeek-AI 2024) have demonstrated remarkable proficiency in natural language understanding, reasoning, and generating structured code. More recently, the trajectory of AI research is shifting toward agentic systems: LLM-powered entities that can plan multi-step tasks, invoke external tools, and maintain internal state over long horizons. For example, recent work has begun to explore LLM-powered agents for structural analysis, such as a multi-agent system for 2D frame analysis that decomposes tasks into problem analysis, geometry derivation, and OpenSeesPy code generation, achieving over 80% modeling accuracy (Geng et al. 2025). Similarly, the MASSE framework assigns distinct agents to interpret design codes, perform load calculations, and verify capacity, automating a real-world design scenario that normally took hours into minutes (Liang et al. 2025). Moreover, agentic frameworks like ALL-FEM have fine-tuned LLMs on verified finite-element code to generate and debug FEniCS simulation scripts across multi-physics problems (Deotale et al. 2026).

However, a common pattern in these systems is an end-to-end or "one-shot" operation: the engineer provides a prompt, and the agents output a complete model or script, often without explicit mechanisms for iterative correction. This approach exposes critical risks, as LLMs are known to hallucinate, make numerical mistakes, and lack awareness of variational structures, meaning errors can go unnoticed without human oversight (Deotale et al. 2026).

Moreover, finite element modeling inherently requires iterative validation, especially for geometry and topology, which depends on visual inspection. Current vision-language models remain insufficient for reliably performing such validation (Zhang et al. 2026). As a result, fully autonomous systems lack mechanisms for systematic error detection, correction, and adaptation (Avila et al. 2025). Research in different engineering domains confirms that incorporating humans into the AI loop mitigates these issues: human-in-the-loop decoding for code generation led to significantly fewer vulnerabilities (González et al. 2025), and digital twin frameworks in civil infrastructure have begun formalizing human-AI collaboration for iterative sensing and control (Sun et al. 2026). These studies emphasize that expert-in-the-loop workflows require the systematic integration of domain expertise throughout the entire process, not merely occasional validation.

To address these limitations in FEA, this paper proposes the Human-Enhanced Loop Modeling (HELM) framework, a collaborative human-AI protocol for finite element modeling. HELM decomposes the modeling workflow into three primary components, i.e., geometry generation, boundary condition definition, and material assignment, handled by specialized agents. Unlike end-to-end systems, HELM introduces structured checkpoints at which intermediate results are presented to the human user for validation and refinement. This human-in-the-loop mechanism enables iterative correction, reduces the risk of propagated errors, and aligns the modeling process with the inherently iterative nature of engineering practice. Conceptually, HELM implements a “skill-rule-knowledge” hierarchy: AI agents manage routine, skill-level operations (e.g., geometry generation) and rule-based tasks (e.g., standard boundary condition assignment), while human experts reserve knowledge-based decisions, like validating intermediate results and resolving novel modeling issues, for higher-level oversight.

The framework is demonstrated through a detailed case study of reinforced concrete (RC) bridge barrier modeling, integrating lightweight LLMs with commercial FE software, including ANSYS (Ansys 2024) and LS-PrePost (LSTC 2019). Notably, the development of agentic systems interfacing with commercial finite element platforms remains rare in the current literature. To facilitate broader adoption and future research, the HELM architecture is open-sourced, providing a flexible foundation that can be adapted to other modeling scenarios.

## Human-Enhanced Loop Modeling Framework

The proposed HELM framework establishes a collaborative protocol between human modelers and AI agents to automate the FE modeling process. As illustrated in Fig. 1, the framework decomposes the complex modeling workflow into three primary tasks: (1) geometry creation and meshing, (2) boundary condition definition, and (3) material property assignment. To execute these tasks, three specialized AI agents were developed: Agent_Geo for geometry and meshing, Agent_BC for constraints and load application, and Agent_Mat for material definitions. The framework’s generality was evaluated by deploying the agents across two distinct FE software environments: Agent_Geo operates within ANSYS (Ansys 2024), while Agent_BC and Agent_Mat function in LS-PrePost (LSTC 2019).

The core innovation of HELM is its human-in-the-loop validation protocol, designed to mitigate the risk of LLM hallucinations in code generation. Each modeling task is segmented into discrete checkpoints, as described in Table. 1 and detailed in Table. A1 (Appendix A). At each checkpoint, the human modeler provides a natural language instruction, and the agent generates an actionable script, i.e., APDL for ANSYS and CFILE/Keyword scripts for LS-PrePost (Fig. 1). Specifically, APDL serves as a parametric language to automate modeling and analysis in ANSYS, while CFILE and Keyword scripts are utilized to execute interface commands and define input data structures in LS-PrePost (Ansys 2025; LSTC 2002, 2010). The human modeler then visually verifies the generated result from the FE software. If the agent’s output is incorrect, the modeler provides feedback, initiating a refinement loop. This approach was essential because a majority of FE modeling tasks require visual validation of geometrical and physical correctness, a capability for which current Vision-Language Models (VLMs) remain insufficient and unreliable (Fu et al. 2024; Lu et al. 2024; Yue et al. 2024). While current checkpoint validations require human input, this framework establishes a foundation for future automated pipelines as more sophisticated AI validation systems are developed.

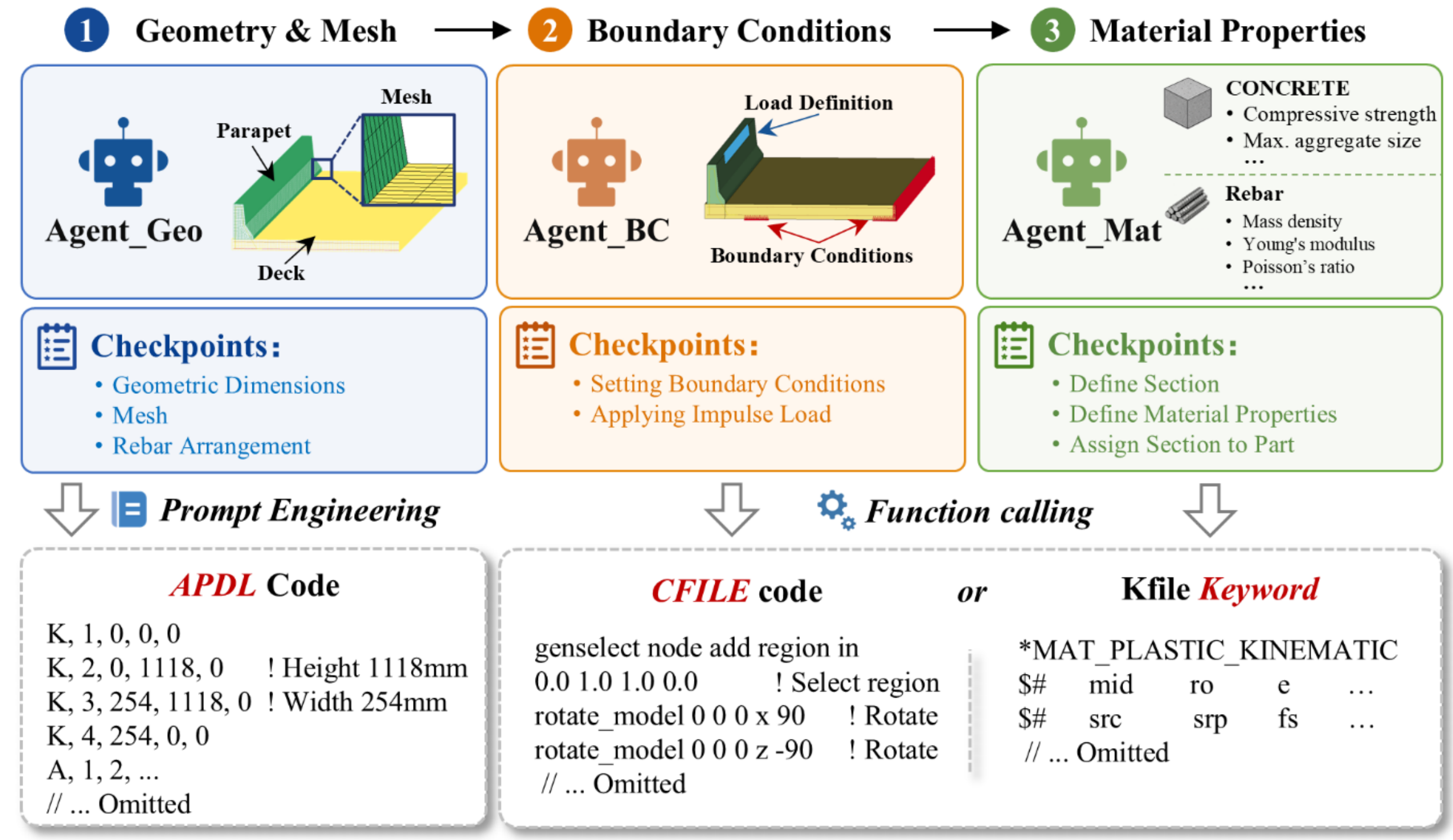

Fig. 1. Illustration of the human-enhanced loop modeling framework

Table 1. Checkpoint Designs for 3D RC barrier Modeling

| Agent Phase | Focus | Action Description | Core Commands |
|---|---|---|---|
| Agent_Geo | Geometry & Mesh | Construct cross-sectional geometry for the parapet and deck, execute 3D solid extrusion, and perform mesh discretization | APDL: K, A, L, LMESH |
| | Rebar Modeling | Establish longitudinal, stirrup, and transverse rebar models; implement longitudinal/lateral distribution and local reinforcement densification | APDL: LSEL, LGEN, LATT |
| Agent_BC | Coordinate Normalization | Execute model rotation and translation to align with the global coordinate system, establishing a spatial reference for automated node identification | .cfile: rotate, translate_model |
| | Constraint Definition | Extract node sets from three support interfaces based on spatial coordinates and apply 6-DOF (Degrees of Freedom) fixed constraints | .k: *SET_NODE_LIST, *BOUNDARY_SPC_SET |
| | Dynamic Load Application | Identify loading points via spatial coordinates, define the AASHTO impulse load curve, and impose transverse dynamic loads | .k: *DEFINE_CURVE, *LOAD_NODE_SET |
| Agent_Mat | Material & Section | Define physical constitutive properties for concrete and steel, and configure sectional physical parameters | .k: *MAT, *SECTION |
| | Property Settings | Associate Material IDs (MID) and Section IDs (SECID) with specific Part IDs (PID) | .cfile: *PART |

To convert human instructions into software-readable scripts, the three agents were constructed using a combination of few-shot prompting and function calling. Few-shot prompting guides the LLM's reasoning and output formatting by providing exemplars of (instruction, script) pairs, which significantly enhances generation reliability over zero-shot approaches (Brown et al. 2020). On the other hand, function calling allows the agents to invoke dedicated, predefined software APIs rather than generating raw script from scratch, improving accuracy and reducing syntax errors (Schick et al. 2023). These techniques create a constrained generation environment, transforming a general-purpose LLM into a domain-specific modeling assistant that can operate commercial computer-aided engineering (CAE) software. The complete source code and designed prompts are accessible at the provided repository [https://github.com/SimAgentDev/Ansys-LSPP-AgentKit].

The framework's performance was demonstrated through a case study of RC bridge barrier modeling (Fig. 2), which encompassed the integrated modeling of the parapet, deck overhang, and internal rebar detailing. In particular, Agent_Geo handles the complete 3D geometry creation and meshing within ANSYS. Subsequently, in LS-PrePost, Agent_BC manages rigid body translations and rotations, physical constraint application, and lateral loading, while Agent_Mat defines the constitutive properties for concrete and reinforcing steel. This case study intends to validate the HELM framework's ability to orchestrate a multi-software, agent-driven modeling process under human supervision.

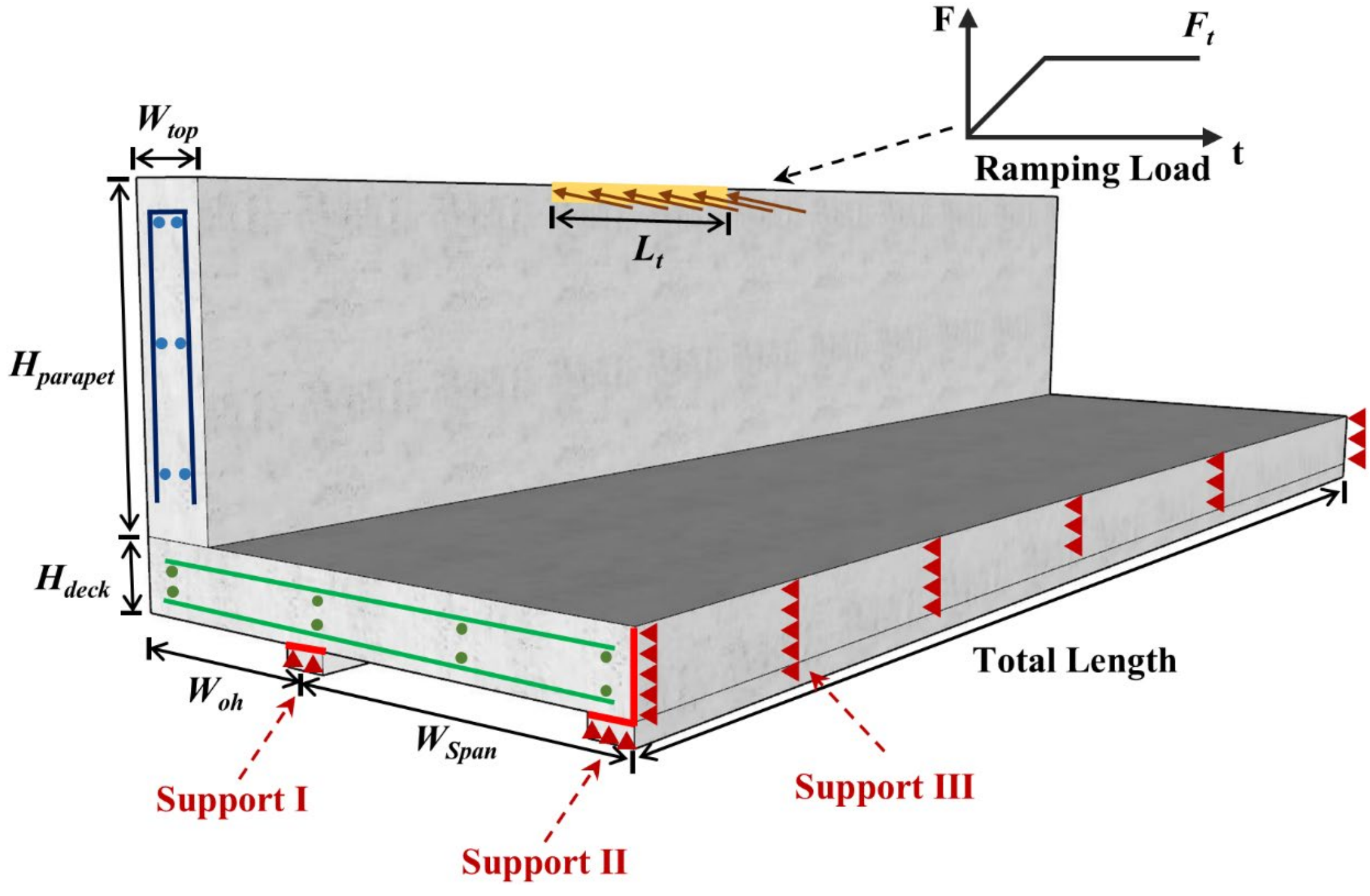

Fig. 2. Overview of the barrier-deck detailing in the case study

## Experimental Designs

The experimental program was designed to evaluate the robustness and generalizability of the proposed HELM framework across a spectrum of realistic bridge barrier configurations. A systematic matrix of 20 representative cases was established, encompassing two MASH Test Levels (TL-4 and TL-5) and three distinct barrier shapes: single slope (SS), New Jersey (NJ), and vertical wall (VW). The design cases were selected from actual bridge barrier design drawings (Alberson et al. 1996, 2004, 2005; Bielenberg et al. 2020; Bullard et al. 2010; Buth 2011; DelDOT 2021; FDOT 2012; Frosch and Morel 2016; Hirsch 1978; Hirsch et al. 1984; Polivka et al. 2005, 2006; Rosenbaugh et al. 2016, 2021; Sheikh et al. 2011). Fig. 3 shows three examples of the barrier-deck systems.

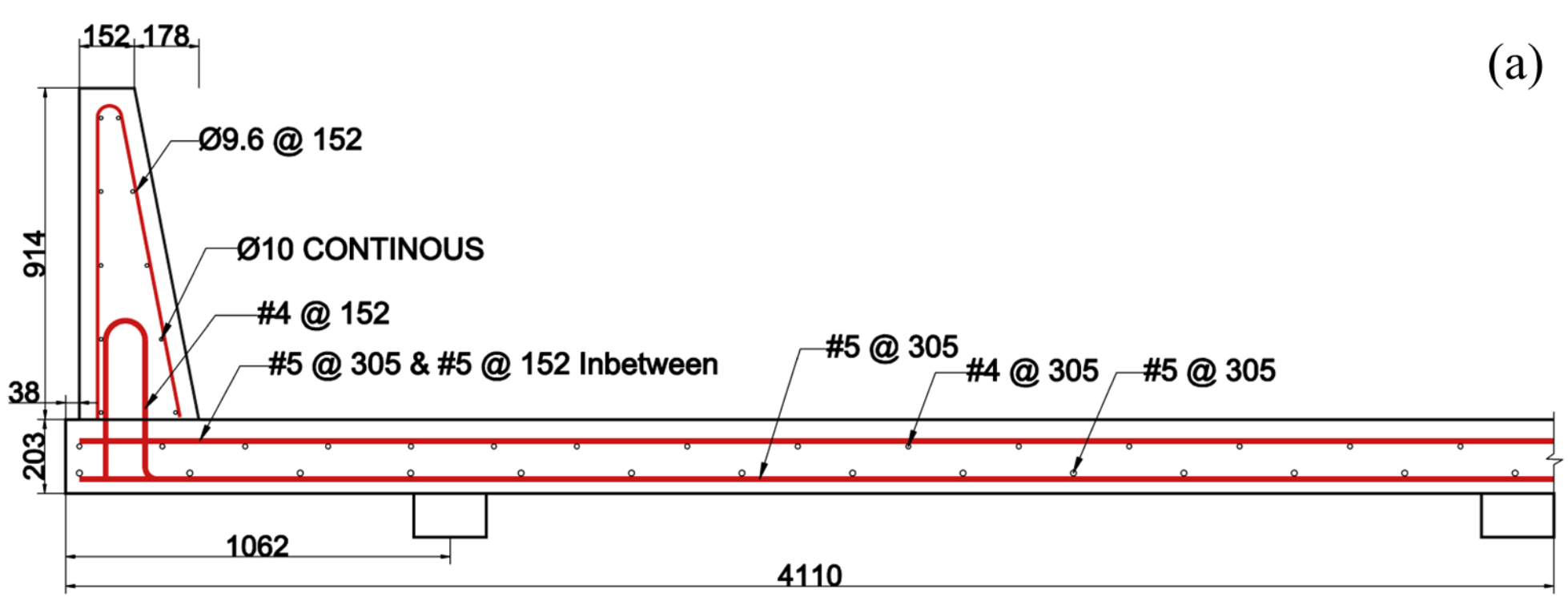

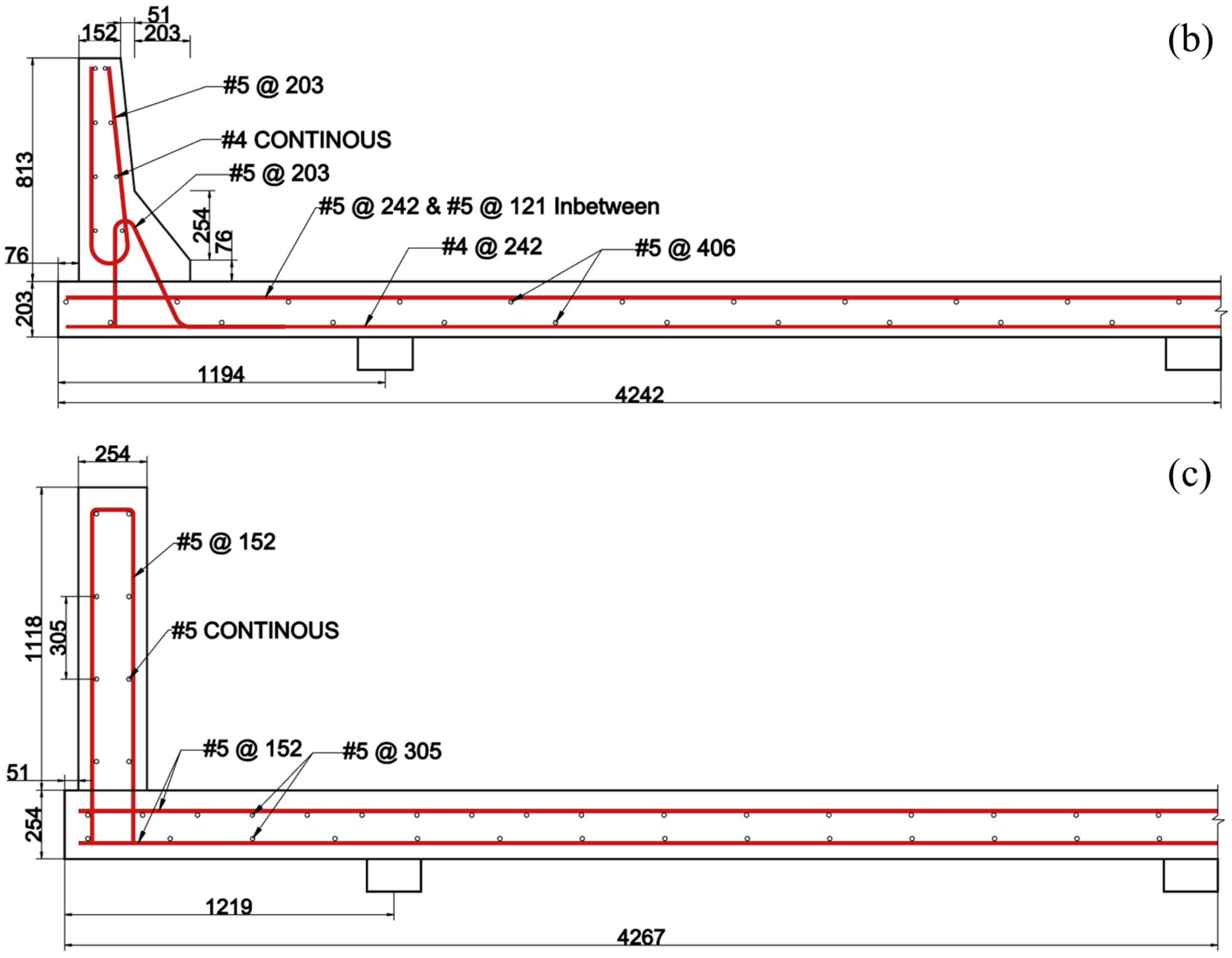

Fig. 3. Geometric detailing and cross-sections of three representative bridge barrier cases (unit: mm): (a) single slope (SS), (b) New Jersey (NJ), and (c) vertical wall (VW)

In alignment with the AASHTO-LRFD specifications, the structural response was evaluated under the most critical loading condition: a quasi-static ramping load applied at the top mid-span of the parapet. The design peak load was distributed over a length of 1,067 mm for TL-4 (240.20 kN) and 2,438 mm for TL-5 (551.58 kN). Material properties were assigned based on case-specific requirements, with concrete grades ranging from C30 to C55 and assuming reinforcement yield strength of 414 MPa. These were implemented in LS-PrePost (LSTC 2019) using material types #159 and #3 (Hallquist 2006). More details about the twenty barrier design cases can be found in Appendix A (Table. A2). The boundary condition of the overhang deck followed the simplified method used in Frosch and Morel (2016) and Cao et al. (2020), where constraints were added to the exterior girders and the inner girders to simulate the performance of barrier-overhang system under lateral loading.

The evaluation protocol for the HELM framework's autonomous modeling capability was structured around a sequence of 22 predefined checkpoints (shown in Appendix A Table. A1), with a rigorous dual-verification criterion at each stage. A checkpoint was considered successful only if it satisfied two conditions: first, the agent-generated instructions must execute successfully within the simulation environment without any syntactic or system errors; second, the resulting geometric topology, component dimensions, and assembly relationships must demonstrate strict adherence to the original engineering design intent. To mitigate the effects of output stochasticity and ensure reliable validation, majority voting across ten independent trials was implemented for each checkpoint, with DeepSeek-V3 (DeepSeek-AI 2024) employed to select the most representative response from the generated ones.

The HELM framework is powered by the Llama-3.1-70B-Instruct-Turbo model (Together AI 2024). This specific model was selected based on its lightweight nature, strong performance in prior structural analysis benchmarks (Geng et al. 2025; Liu et al. 2026), and ease of local implementation. In the safety-critical domain of civil engineering, data privacy and model security are paramount. Therefore, by utilizing a lightweight and open-sourced model that can be hosted locally, the HELM framework mitigates the risks associated with information leakage and prompt injection attacks that target proprietary engineering data.

To assess the framework's capacity for self-correction, a human-assisted retry mechanism was introduced whenever the baseline execution failed. As shown in Fig. 4, once failed the first trial in the checkpoint, the agent would be provided with the specific error message and allowed a single re-attempt to reason and correct its output. If the retry also failed, the ground-truth code will be executed to advance the process to the next checkpoint, thereby preserving the independent evaluation of subsequent agents. The overall performance was rigorously quantified by calculating success rates at two levels: a case-level success, defined by the completion of all 22 checkpoints, and an agent-level success rate, which independently assessed the checkpoints corresponding to each of the three specialized agents, i.e., geometry, boundary condition, and material settings.

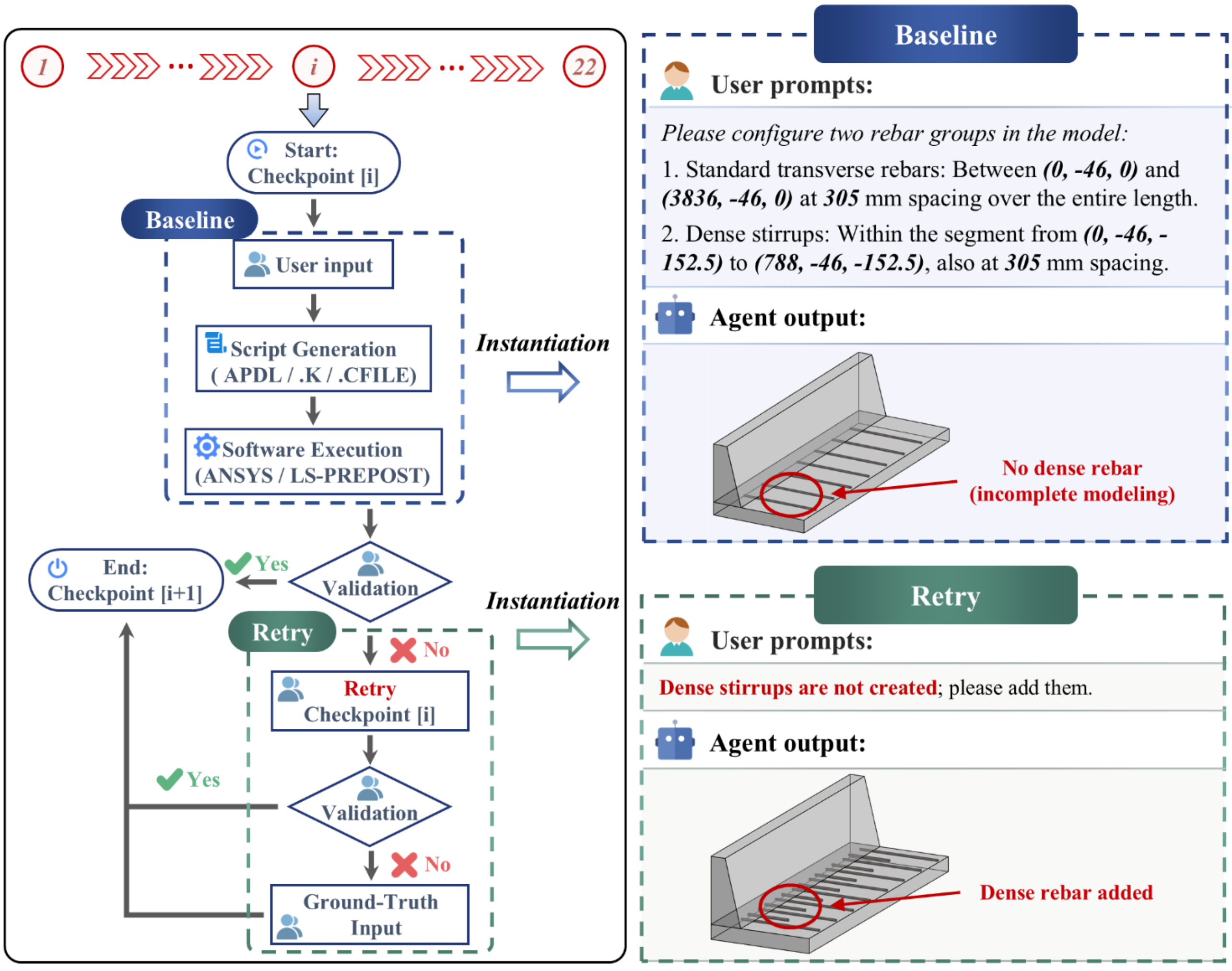

Fig. 4. Flowchart of the checkpoint evaluation

## Evaluation Results and Discussion

As illustrated in Fig. 5, the proposed HELM framework substantially improves the reliability of the agentic modeling system. The baseline configuration, without human-in-the-loop retry, achieves an overall pass rate of 20% across 20 modeling cases. With the introduction of the retry mechanism, the pass rate increases to 75%, clearly demonstrating the effectiveness of incorporating human feedback. At the agent level, the baseline pass rates for Agent_Geo and Agent_BC are 40% and 45%, respectively, which improve to 80% and 90% under HELM. In contrast, Agent_Mat achieves 100% accuracy in the baseline setting and no retry mechanism was involved.

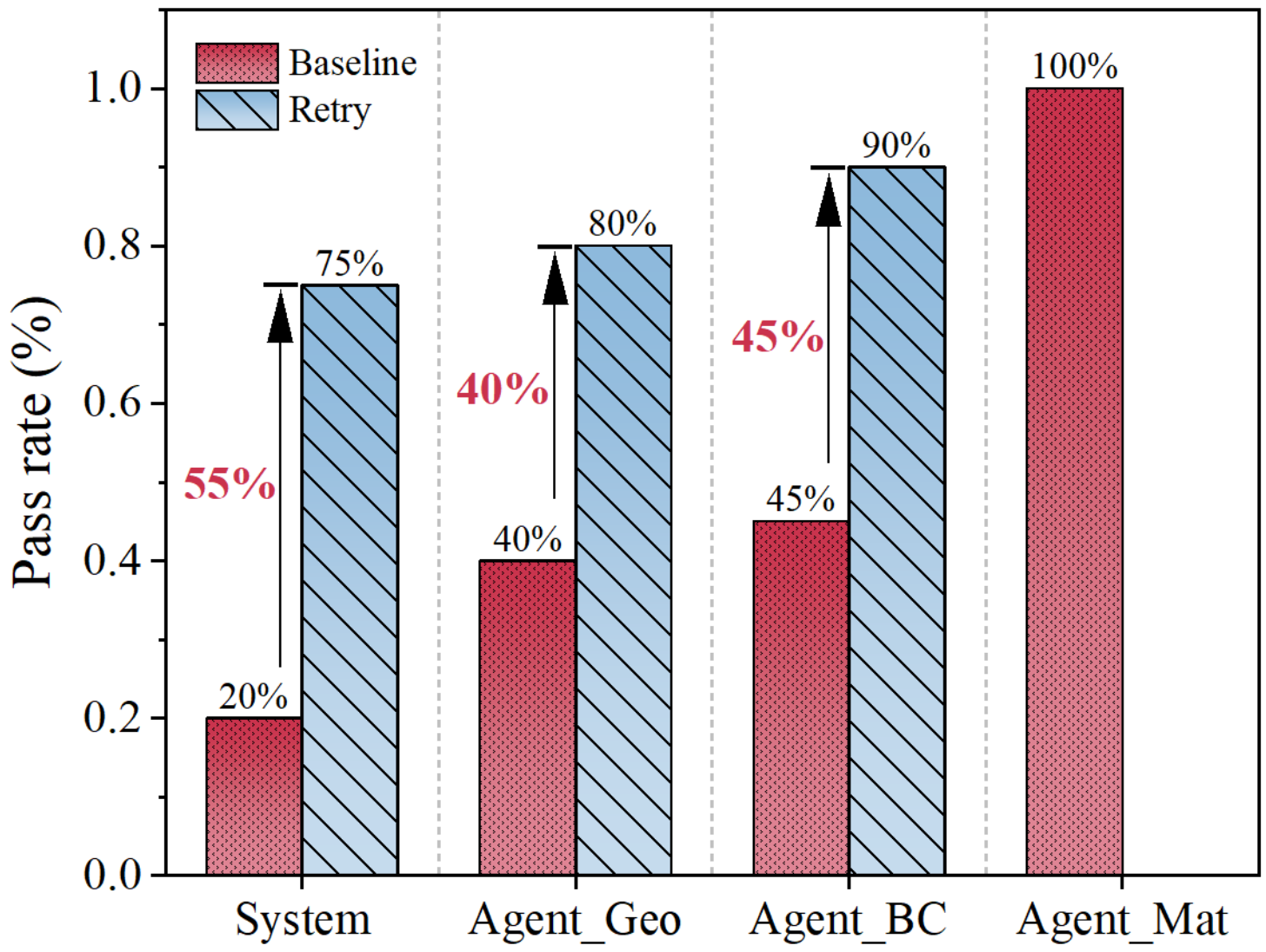

Fig. 5. The overall pass rate and evaluation results of three agents

This performance disparity highlights the fundamentally different levels of difficulty across modeling tasks. Geometry construction and boundary condition definition require handling complex topologies and spatial constraints, placing significantly higher demands on reasoning capabilities. Under challenging scenarios, such as non-uniform reinforcement densification and spatial coordinate-based nodal selection, the initial pass rates of Agent_Geo and Agent_BC fall below 50%. However, with the intervention enabled by HELM, their performance improves markedly, with success rates approximately doubling. This consistent improvement indicates that the HELM framework effectively stabilizes the multi-agent system and guides it toward correct solutions under complex engineering constraints.

Another key advantage of HELM lies in its checkpoint-based design, which decomposes long-horizon modeling tasks into verifiable atomic steps under human scrutiny. This structure enables precise localization of failures and facilitates efficient correction. Notably, even when human supervisors provide only high-level or intuitive feedback (e.g., identifying geometric inconsistencies), the system can leverage localized contextual constraints at checkpoints to autonomously refine subsequent actions. This collaborative paradigm enhances robustness without requiring sophisticated human instructions, thereby maintaining both efficiency and scalability. It should also be noted that without the agentic system design, general foundation models with larger sizes, like DeepSeek-V3.2 (DeepSeek-AI 2024) and Qwen3.5 (Qwen Team 2026), could not finish a single case of the modeling task (as shown in Fig. A2 along with the testing prompts).

Despite the significant improvements achieved by HELM, failure cases still occur, particularly in complex modeling scenarios. A total of 31 failure cases were identified and their distributions are summarized in Table 2, with representative examples shown in Fig. 6.

Table 2. Modeling error types at the checkpoint level

| Specific Error Type | Occurrences |
| --- | --- |
| Geometric Modeling Errors | 14 |
| Confusion in Algebraic Logic | 13 |
| Data Type Mismatch | 2 |
| Calculation Error | 2 |
| Total | 31 |

## (a) Geometric modeling errors

**Ground truth** & **Erroneous Result:**

```
K, KMAX+1, -19.6, -59, -121
K, KMAX+2, 1461, -59, -121
L, KMAX+3, KMAX+4
*GET, LMAX_DENSE, LINE, , NUM, MAX
LSEL, S, LINE, , LMAX_DENSE
*SET, DENSE_LINE, LMAX_DENSE
ET, 14, BEAM161
LATT, 1, 1, 14, , 111, , 1
LESIZE, ALL, 1.5*25.4, ,
LMESH, ALL
LSEL, S, LINE, , TRANSVERSE_LINE
*SET, ITIME_TRANSVERSE, 63
LGEN, ITIME_TRANSVERSE, ALL, , , , , -242, , ,
LSEL, S, LINE, , DENSE_LINE
*SET, ITIME_DENSE, 62
LGEN, ITIME_DENSE, ALL, , , , , -242, , ,
ALLSEL, ALL
```

! Omitted

! Omitted

**Ground truth:**

Parapet

Transverse rebar

Omission of dense reinforcement

Deck

**Ground truth** & **Erroneous Result:**

```
K, KMAX+1, 76.2, 30, 0 ! Point 1 (Z=0 mm)
K, KMAX+2, 76.2, -144, 0 ! Point 2
K, KMAX+3, 371.6, -144, 0 ! Point 3
K, KMAX+4, 371.6, 59.4, 0 ! Point 4
K, KMAX+5, 113.5, 439.8, 0 ! Point 5
L, KMAX+1, KMAX+2 ! Point 1-Point 2 (straight line)
K, KMAX+6, ((76.2+371.6)/2)+(ABS(371.6-76.2))*0.1,
((-144+59.4)/2)+(ABS(371.6-76.2))*0.4, 0
LARC,  KMAX+2,  KMAX+3,  KMAX+6
L, KMAX+3, KMAX+4 ! Point 3-Point 4 (straight line)
L, KMAX+4, KMAX+5 ! Point 4-Point 5 (straight line)
```

! Redundancy

**Ground truth:**

**Erroneous Result:**

Incorrect Shape

## (b) Confusion in Algebraic Logic

**User: Select nodes at -2057.0 < Y < -839.0, where Z =0.0**

**Erroneous Result:**

Call: Select_Nodes (y_range = *[-839,  -2057]*)  ! Inverted values

**Ground truth:**

Call: Select_Nodes (y_range = *[-2057,  -839]*)

## (c) Data Type Mismatch

**User: Initialize model direction**

**Erroneous Result:**

LLM Assistant: Calling tool 'normalize_model_direction',
arguments: (// ... Omitted
"min_x": "***null***", "max_y": "None", "min_z: "None")   ! TypeError

**Ground truth:**

LLM Assistant: ( ... "min_x": "*None*"...)

## (d) Calculation Errors

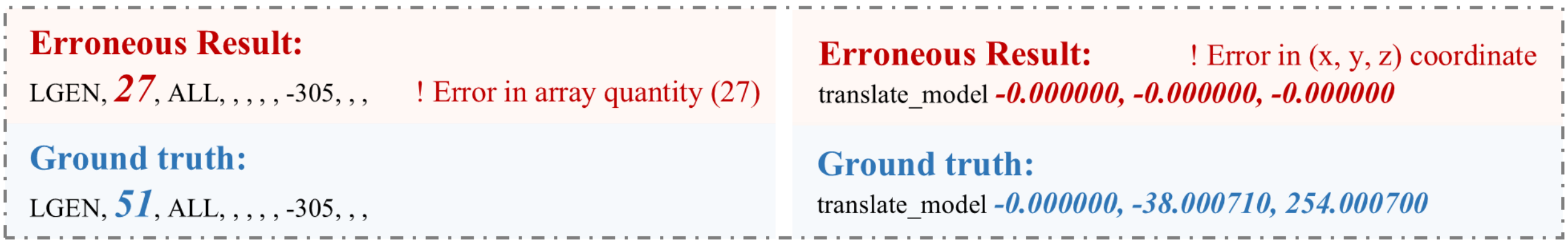

Fig. 6. Description of the error types

(a) Geometric Modeling Errors

Geometric errors constitute the most prevalent failure mode, primarily arising from insufficient spatial reasoning capabilities. These errors manifest as omissions in complex topological definitions and inconsistencies in geometric configurations. As shown in Fig. 6a, the system often fails to model reinforcement layouts involving locally densified spacing, resulting in incomplete structural representations. Additionally, the model may misinterpret local geometric features, such as confusing straight segments with arc-based stirrup geometries.

(b) Confusion in Algebraic Logic

A second major category involves logical inconsistencies in handling numerical and spatial constraints. These errors typically appear in interval reasoning and ordering relationships, such as reversing node selection ranges (e.g., changing [-2057, -839] into [-839, -2057]). As a result, generated instructions may be syntactically valid but physically incorrect. This issue indicated that integrating external symbolic or numerical reasoning tools could be necessary for reliable deployment in engineering applications.

(c) Data Type Mismatch

Data type mismatches arise from improper handling of output formats, such as generating string values (e.g., "null") where numerical types or Python None are required. These inconsistencies lead to interface parsing failures and pipeline interruptions during function calling. Although less frequent, such errors highlight the importance of strict data type enforcement and validation mechanisms when integrating LLMs into engineering workflows.

(d) Calculation Errors

Calculation-related errors are less frequent but still present. These errors stem from inaccuracies in numerical computation rather than syntax generation. For instance, incorrect array dimensions (e.g., fail to calculate the number of rebars using total_length/spacing + 1 = (15240/300)+1 = 51, shown in Fig. 6d) can invalidate downstream processes. Similar numerical inconsistencies have been reported in recent mechanics-oriented LLM benchmarks (e.g., Wan et al. 2025), reinforcing that arithmetic precision remains a limitation of current LLMs.

Overall, the error distribution suggests that the primary limitations of the agentic system are associated with spatial reasoning and algebraic consistency. Compared with existing studies on LLM-based agents in specialized domains, the HELM framework demonstrates a clear advantage in mitigating these issues through structured human feedback and checkpoint-based correction. However, further improvements may require hybrid approaches that integrate LLMs with domain-specific solvers or symbolic reasoning modules to address fundamental limitations in numerical and spatial reasoning.

## Conclusions

This paper introduced the Human-Enhanced Loop Modeling (HELM) framework, a collaborative human-agent protocol designed to address the reliability and validation challenges inherent in autonomous finite element modeling. Focusing on a representative high-stakes case study, reinforced concrete bridge barrier modeling, the HELM framework was demonstrated to decompose long-sequence modeling workflows into discrete, verifiable checkpoints, enabling structured human intervention at critical stages of geometry generation, boundary condition definition, and material assignment.

The study makes three primary contributions. First, it proposes and evaluates a novel CAE interface that formalizes human-agent collaboration, moving beyond end-to-end autonomous generation toward an iterative, human-supervised paradigm aligned with engineering practice. Second, as the first open-sourced agentic framework to interface with both ANSYS APDL and LS-PrePost, HELM provides reusable architectural guidance for two widely used but previously under-supported commercial finite element platforms. Third, experimental results across a 20-case design matrix demonstrate that the HELM framework improves the baseline single-pass modeling success rate from 20% to 75%, with agent-level pass rates for geometry and boundary condition tasks approximately doubling. Agent_Mat achieved 100% baseline accuracy, reflecting the comparatively lower spatial reasoning demands of material property definition. Error analysis revealed that the dominant failure modes, geometric modeling errors and algebraic logic confusion, reflect fundamental limitations in spatial reasoning and symbolic numerical consistency of current large language models. The checkpoint-based human-in-the-loop design of HELM effectively localizes and mitigates these failures, even with limited or high-level human feedback.

More broadly, these findings indicate that the primary barriers to agentic CAE adoption in performance-based structural design lie not in code generation but in robust spatial reasoning and algebraic verification. The HELM framework demonstrates that structured human oversight can substantially close this reliability gap. Future work should explore the integration of external symbolic solvers, geometric reasoning modules, and computer vision tools as automated validation mechanisms that progressively reduce the human effort required at each checkpoint, moving toward a more autonomous yet equally reliable modeling pipeline.

## Appendix A: Supplementary Checkpoint Specifications, Case Configurations, and Evaluation Results

Table A1. Checkpoint Specifications for Parapet Modeling

| Agent | Checkpoint | Operation | Description |
|---|---|---|---|
| Agent_Geo (ANSYS) | 1 | Parapet Cross-Section Definition | Delineate the parapet cross-sectional geometry by interconnecting vertices at user-specified coordinates |
| | 2 | Deck Cross-Section Definition | Delineate the deck cross-sectional geometry by interconnecting vertices at user-specified coordinates |
| | 3 | Longitudinal Geometric Extrusion & Meshing | Execute 3D solid generation through profile extrusion, followed by mesh discretization based on prescribed element sizes |
| | 4 | Parapet Longitudinal Rebar Modeling | Model the longitudinal rebars and distribute them at specified intervals along the barrier |
| | 5 | Parapet Stirrup Modeling | Define the stirrup profiles and arrange them at a set spacing along the longitudinal axis |
| | 6 | Deck Top Transverse Rebar Modeling | Define the top transverse rebar and implement longitudinal distribution along the bridge axis based on specified spacing, including adaptive densification in critical reinforcement zones |
| | 7 | Deck Bottom Transverse Rebar Modeling | Define the bottom transverse rebar and implement longitudinal distribution along the bridge axis based on specified spacing |
| | 8 | Deck Top Longitudinal Rebar Modeling | Define the top longitudinal rebar and implement lateral distribution across the deck width based on specified spacing |
| | 9 | Deck Bottom Longitudinal Rebar Modeling | Define the bottom longitudinal rebar and implement lateral distribution across the deck width based on specified spacing |
| Agent_BC (LS-PrePost) | 10 | Standardized Orientation Alignment | Align the model's orientation with the prescribed global coordinate system to establish a spatial reference for subsequent automated node identification |
| | 11 | Origin Normalization | Translate the model to the global origin (0,0,0) to enable node selection based on spatial coordinates |
| | 12 | Boundary Support Node I Selection | Select nodes of the initial support interface (Support I) based on spatial coordinate to define the Constraint_NodeSet |
| | 13 | Boundary Support Node II Selection | Select nodes of the subsequent support interface (Support II) based on spatial coordinate and append them to the Constraint_NodeSet |
| | 14 | Boundary Support Node III Selection | Select nodes of the longitudinal symmetry plane (Support III) based on spatial coordinate and append these nodes to the Constraint_NodeSet |
| | 15 | Boundary Condition Application | Impose a fixed boundary condition to the Constraint_NodeSet to restrict all 6 DOFs (X, Y, Z, Rx, Ry, Rz) |
| | 16 | Loading Node Selection | Select nodes along the parapet's top edge (per AASHTO-LRFD 2024) based on spatial coordinates and define them as the Loading_NodeSet |
| | 17 | Load Curve Definition | Define the impulse Loading_Curve via discrete (time, force) data points (per AASHTO-LRFD 2024]) |
| | 18 | Dynamic Load Imposition | Impose the specified transverse force onto the Loading_NodeSet via the defined Loading_Curve |
| Agent_Mat (LS-PrePost) | 19 | Concrete Constitutive Modeling | Define the concrete material properties, including the failure pressure limit and the maximum aggregate size |
| | 20 | Steel Constitutive Modeling | Define the steel material properties, including mass density, Young's modulus, and Poisson's ratio |
| | 21 | Dimensional Parameter Setup | Define the physical thickness of the elements to enable the subsequent association between sectional attributes and geometric parts |
| | 22 | Material Property Assignment | Incorporate material identification (MID) and sectional properties (SECID) into specific Part IDs (PID) to finalize the structural characterization of the parapet system |

Table A2. Cross-section and reinforcement details of the barrier

| Case ID | Test Level | Barrier Shape | Drawing |
|---|---|---|---|
| 01 | TL-4 | Single Slope | Material Strength<br>Concrete: C30<br>Rebar: ASTM A615 Gr. 60<br>Unit: mm |
| 02 | TL-4 | Vertical | Material Strength<br>Concrete: C30<br>Rebar: ASTM A615 Gr. 60<br>Unit: mm |
| 03 | TL-4 | New Jersey | Material Strength<br>Concrete: C40<br>Rebar: ASTM A615 Gr. 60<br>Unit: mm |
| 04 | TL-4 | Vertical | Material Strength<br>Concrete: C30<br>Rebar: ASTM A615 Gr. 60<br>Unit: mm |
| 05 | TL-4 | Single Slope | Material Strength<br>Concrete: C30<br>Rebar: ASTM A615 Gr. 60<br>Unit: mm |

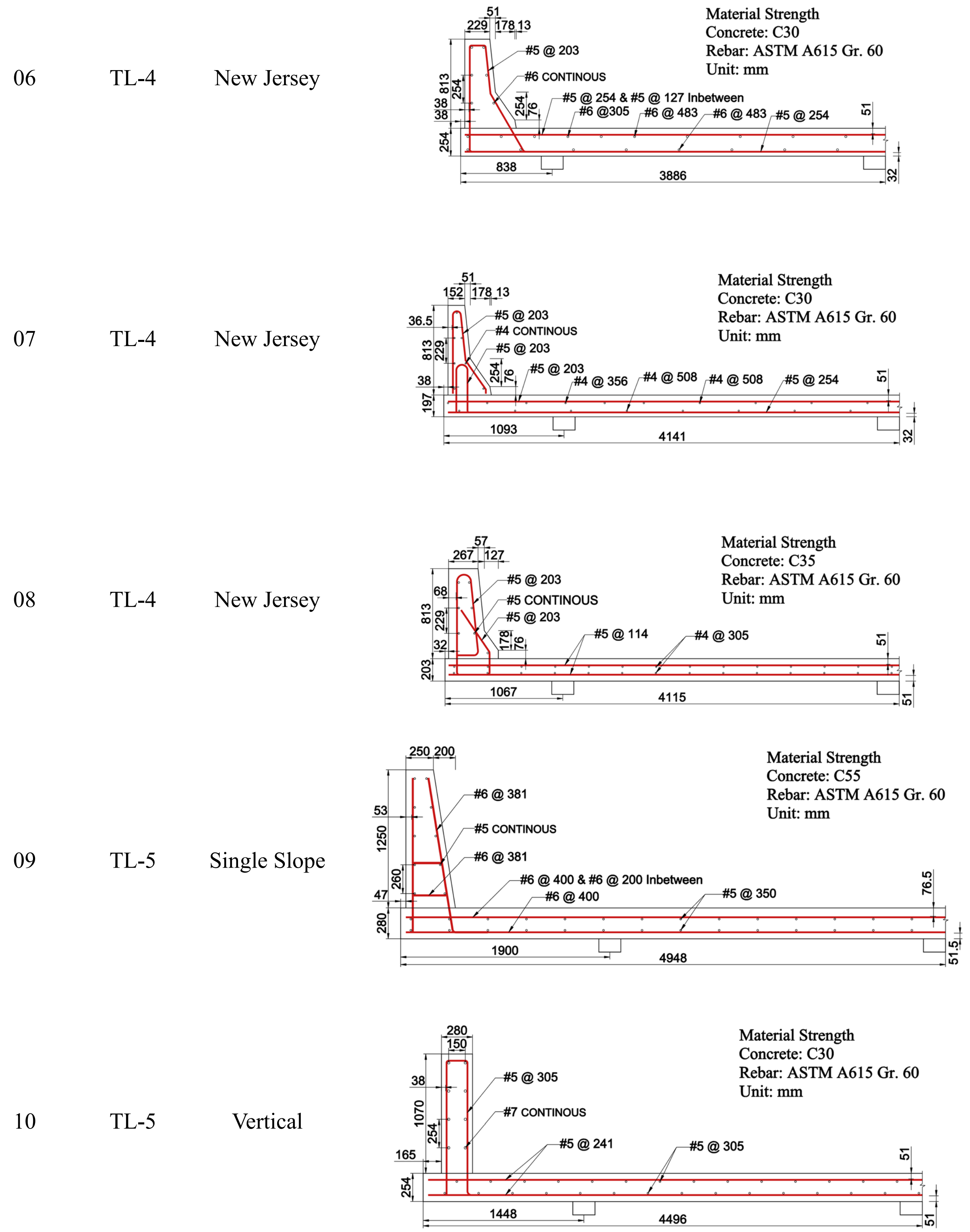
06
TL-4
New Jersey
Material Strength
Concrete: C30
Rebar: ASTM A615 Gr. 60
Unit: mm
#5 @ 203
#6 CONTINOUS
#5 @ 254 & #5 @ 127 Inbetween
#6 @305
#6 @ 483
#6 @ 483
#5 @ 254
838
3886
07
TL-4
New Jersey
Material Strength
Concrete: C30
Rebar: ASTM A615 Gr. 60
Unit: mm
#5 @ 203
#4 CONTINOUS
#5 @ 203
#5 @ 203
#4 @ 356
#4 @ 508
#4 @ 508
#5 @ 254
1093
4141
08
TL-4
New Jersey
Material Strength
Concrete: C35
Rebar: ASTM A615 Gr. 60
Unit: mm
#5 @ 203
#5 CONTINOUS
#5 @ 203
#5 @ 114
#4 @ 305
1067
4115
09
TL-5
Single Slope
Material Strength
Concrete: C55
Rebar: ASTM A615 Gr. 60
Unit: mm
#6 @ 381
#5 CONTINOUS
#6 @ 381
#6 @ 400 & #6 @ 200 Inbetween
#6 @ 400
#5 @ 350
1900
4948
10
TL-5
Vertical
Material Strength
Concrete: C30
Rebar: ASTM A615 Gr. 60
Unit: mm
#5 @ 305
#7 CONTINOUS
#5 @ 241
#5 @ 305
1448
4496

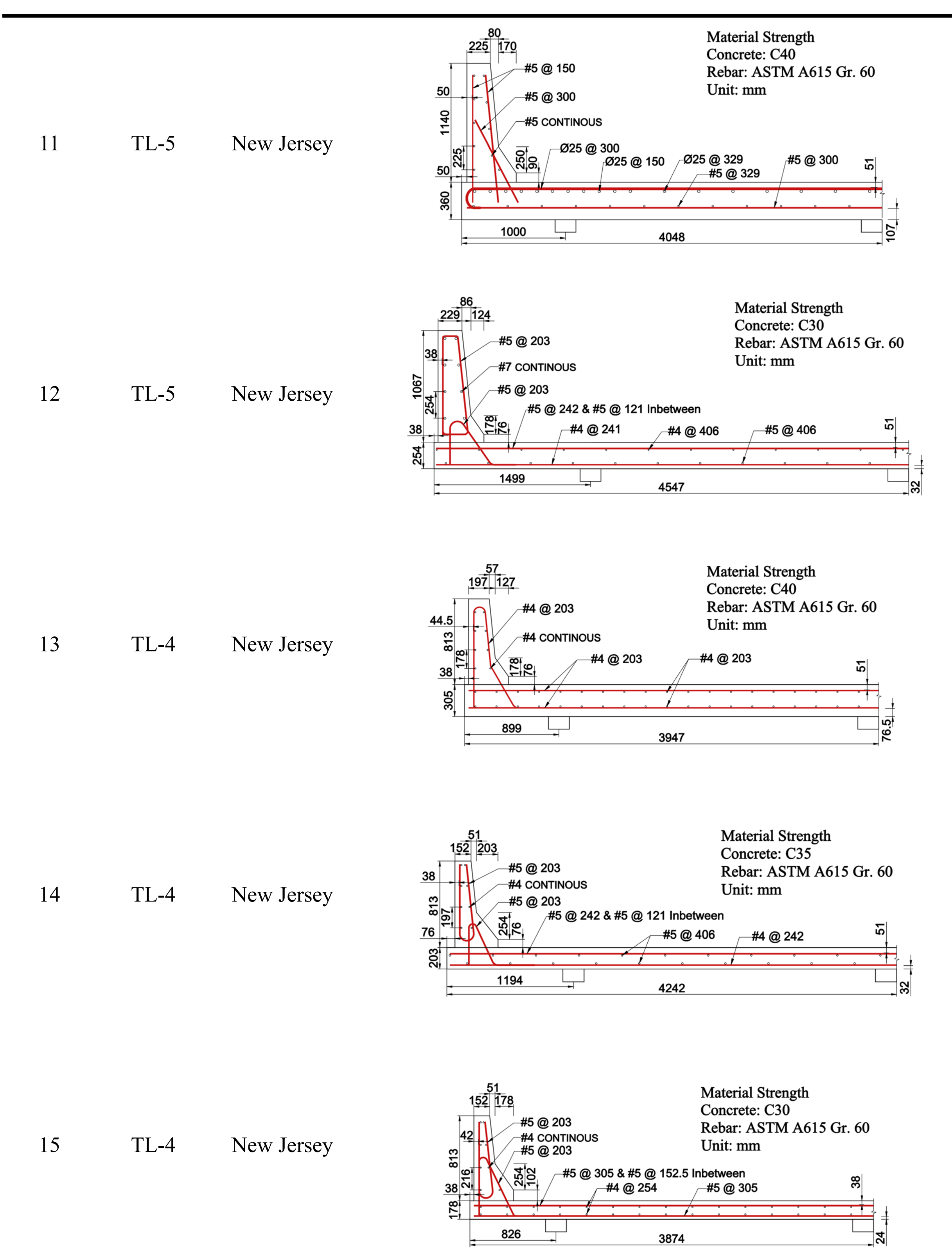

| | | | |
|---|---|---|---|
| 11 | TL-5 | New Jersey | |
| 12 | TL-5 | New Jersey | |
| 13 | TL-4 | New Jersey | |
| 14 | TL-4 | New Jersey | |
| 15 | TL-4 | New Jersey | |
| 16 | TL-4 | Vertical | |
| 17 | TL-4 | Single Slope | |
| 18 | TL-5 | Vertical | |
| 19 | TL-5 | Single Slope | |
| 20 | TL-4 | New Jersey | |

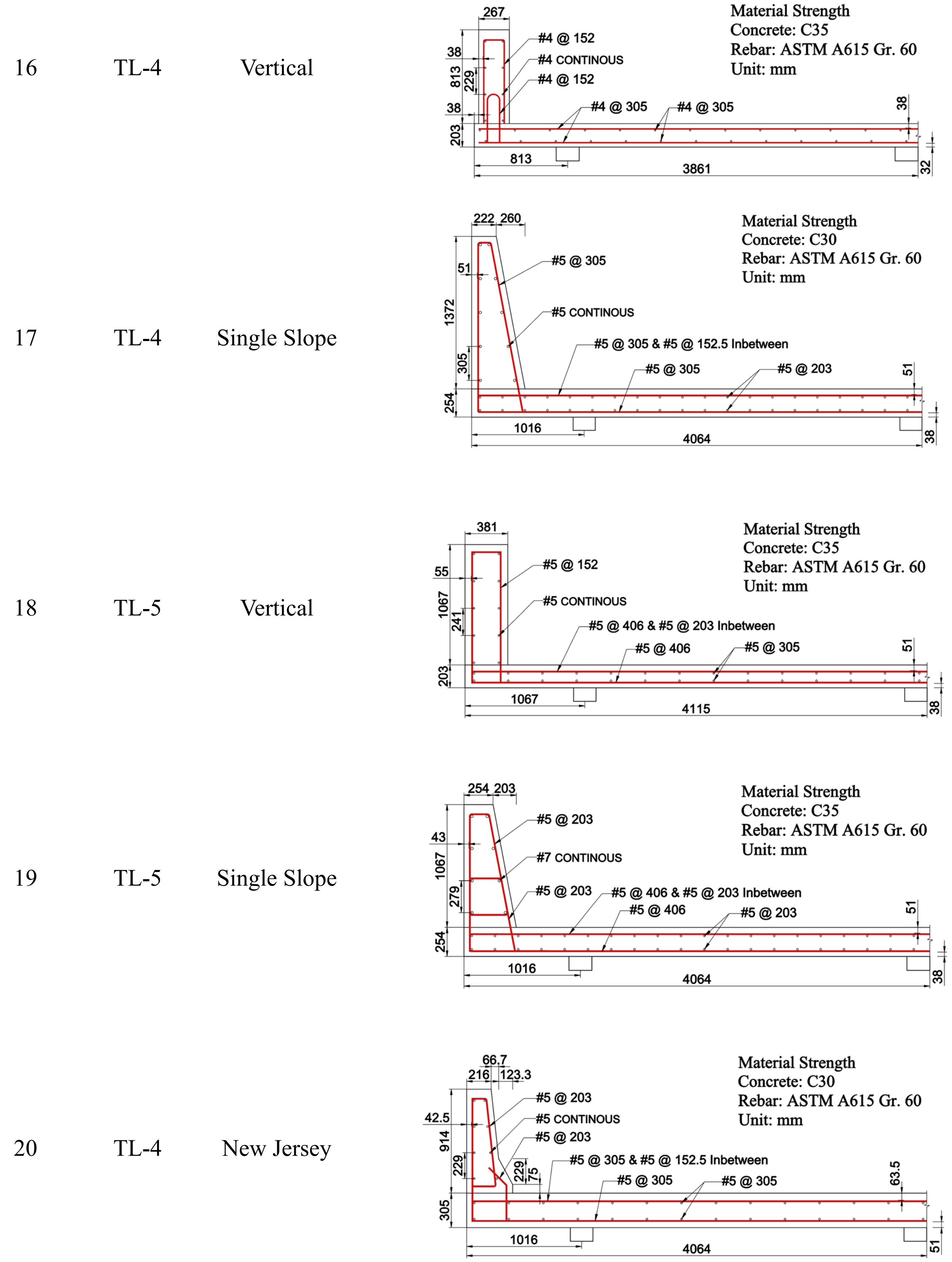

Fig. A1 compares the checkpoint-level accuracy of the baseline and retry configurations across the 22 predefined checkpoints in the HELM framework, with each checkpoint's accuracy averaged over 20 test cases. While most checkpoints achieved perfect accuracy under the baseline setting, several geometry-generation and boundary-condition checkpoints exhibited reduced success rates due to their higher spatial reasoning and coordinate-processing complexity. The introduction of the retry mechanism substantially improved performance across these challenging checkpoints, with several checkpoints achieving complete recovery to 100% accuracy. These results demonstrate the effectiveness of the human-in-the-loop intervention strategy in correcting localized modeling errors and improving the robustness of long-sequence finite element modeling workflows.

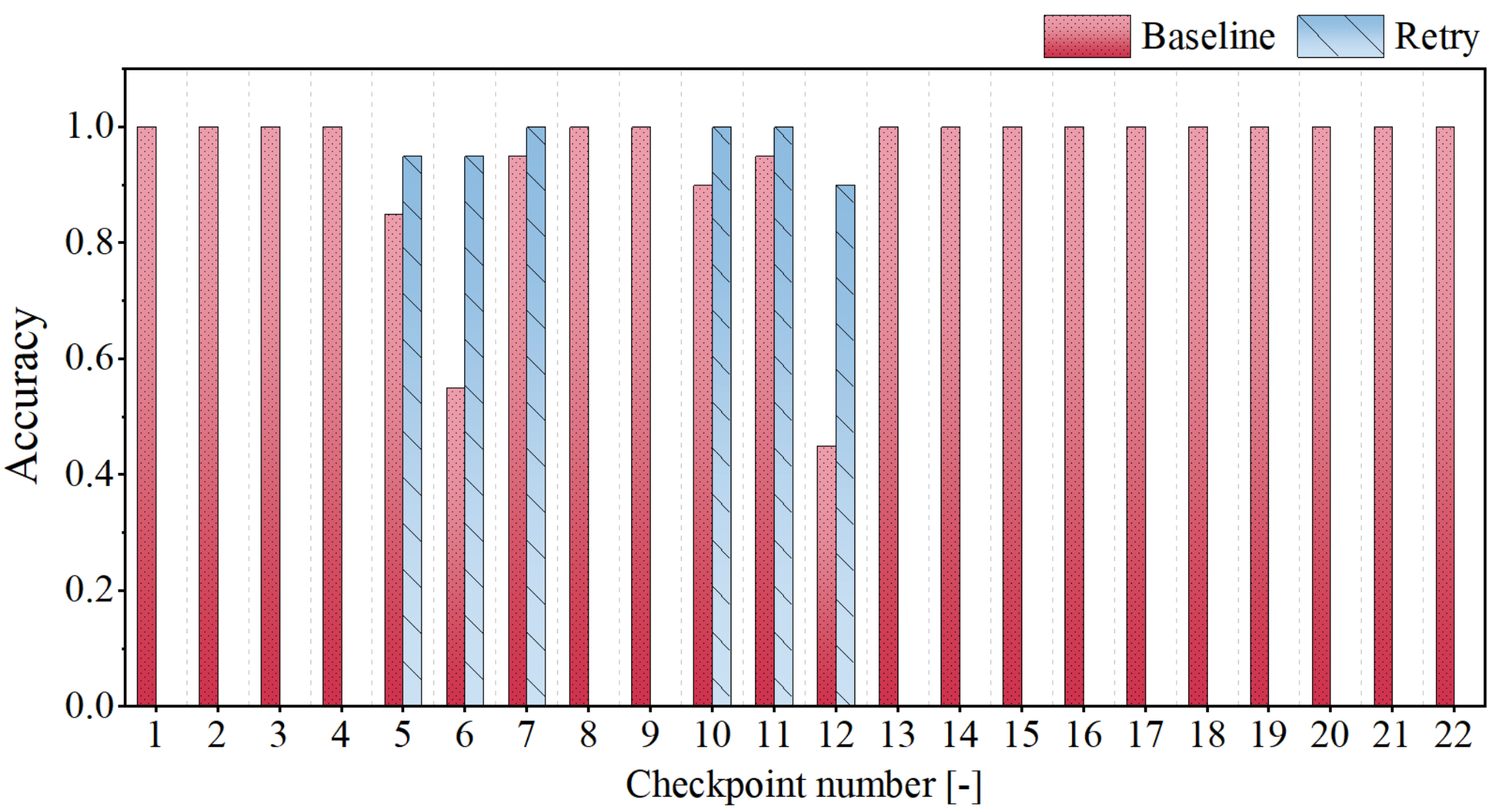

Fig. A1. Accuracy of Baseline and Retry across different checkpoints

Fig. A2 evaluates and compares the performance of different Large Language Model (LLM) implementations within the Agent_Geo phase. All results shown in this figure were achieved through end-to-end implementation. Agent_Geo represents the proposed checkpoint-based agentic workflow, whereas DeepSeek-V3.2 and Qwen3.5 were evaluated using an end-to-end modeling strategy without task decomposition. The results show that Agent_Geo achieved pass rates of 40% and 80% under the baseline and retry settings, respectively. In contrast, both end-to-end LLMs failed to successfully complete any modeling case, yielding pass rates of 0%. These results highlight the importance of task decomposition and iterative verification in long-sequence finite element modeling. Appendix B provides five representative prompting examples used by Agent_Geo, including the system prompts, user prompts, and corresponding modeling instructions.

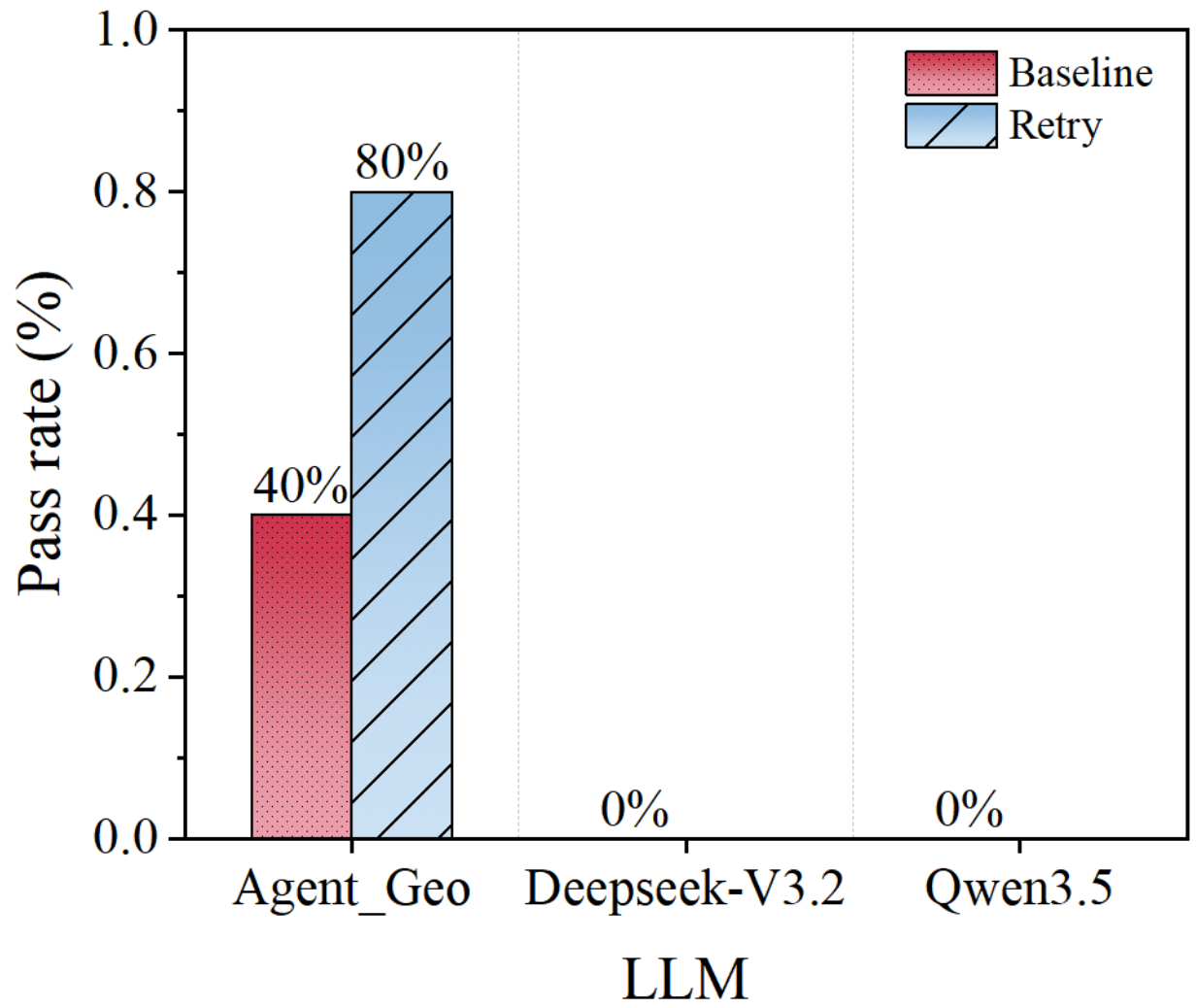

Fig. A2. Pass rate of distinct LLMs

## Appendix B: Representative Prompting Examples and End-to-End APDL Generation by Agent_Geo

Fig. B1 presents the system prompt provided to the LLMs for end-to-end geometry modeling. Figs. B2 and B3 show the user input and the corresponding APDL code generated by DeepSeek-V3.2 for Case 01, respectively. Similarly, Figs. B4–B5, B6–B7, B8–B9, and B10–B11 present the user inputs and generated APDL scripts for Cases 02 through 05.

🛠️ **System Prompt:**

Role: You are an expert APDL Code Generator for barrier modeling.
Task: Generate correct, complete, and executable APDL code based on user descriptions.
Focus: Accurate geometric modeling & reinforcement arrangement.

Fig. B1. Prompt for End-to-End Geometry Modeling

## User Prompt: Barrier & Deck Modeling Specifications

Task: Generate complete APDL code for barrier modeling based on the user description.
Output: Generate the COMPLETE APDL code for the barrier model described above.
Requirements:

- Code must be complete, executable, and free of placeholders.
- Include all necessary engineering modeling steps.
- Apply strict, proper APDL native commands and syntax.
- Include descriptive comments explaining key operations.
- Ensure the entire script is organized, indented, and readable.

**I. Basic Modeling Parameters**

Coordinate origin (0, 0, 0): Intersection of the vertical back face of the barrier and the top surface of the bridge deck.
Unit: millimeters (mm).
Total length (L): 15240 mm (model extends in the negative Z-axis direction).

**II. Solid Modeling**

1. Barrier Cross-Section Profile

Connect the following vertices in sequence to form a closed polyline, then extrude it 15240 mm along the Z-axis:

- P1: (0, 0, 0)
- P2: (0, 991, 0)
- P3: (203, 991, 0)
- P4: (254, 0, 0)

2. Deck Cross-Section Profile

Connect the following vertices in sequence to form a closed polyline, then extrude it 15240 mm along the Z-axis:

- D1: (-51, 0, 0)
- D2: (-51, -203, 0)
- D3: (5283, -203, 0)
- D4: (5283, 0, 0)

**III. Barrier Reinforcement**

1. Longitudinal Rebars

All longitudinal rebars have a length of 15240 mm (extending from Z=0 to Z=−15240).

- Left side (vertical face) rebar array:
  Initial rebar: from (71.5, 222, 0) to (71.5, 222, -15240)
  Copy logic: Copy upward (positive Y-axis) 3 times at 222 mm spacing.
- Right side (sloped face) rebar array:
  Positioning: Take the top rebar on the left side, copy it rightward (positive X-axis) once at 65.23 mm spacing to obtain the starting rebar.
  Copy logic: Copy this starting rebar downward and rightward 3 times; displacement increment: 11.42 mm horizontally right, 222 mm vertically downward.

2. Barrier Stirrups

Single stirrup path points:

- (57, -179, 0)
- (57, 888, 0)
- (151.2, 888.7, 0) — Note: An upward convex arc shall be created between P2 and P3
- (206.1, -179, 0)

Distribution: Array along the negative Z-axis at 305 mm spacing to cover the full length.

**IV. Deck Reinforcement**

1. Transverse Rebars

- Left side (vertical face) rebar array:
  Initial rebar: from (71.5, 222, 0) to (71.5, 222, -15240)
  Copy logic: Copy upward (positive Y-axis) 3 times at 222 mm spacing.
- Right side (sloped face) rebar array:
  Positioning: Take the top rebar on the left side, copy it rightward (positive X-axis) once at 65.23 mm spacing to obtain the starting rebar.
- Upper transverse rebars: Straight line connecting (33.5, -79, 0) and (5283, -79, 0).
- Lower transverse rebars: Straight line connecting (34.6, -179, 0) and (5283, -179, 0).

Distribution: Both layers arrayed along the negative Z-axis at 305 mm spacing.

2. Longitudinal Rebars

All longitudinal rebars have a length of 15240 mm (extending from Z=0 to Z=−15240).

- Upper longitudinal rebars:
  Initial rebar: from (216.4, -93.5, 0) to (216.4, -93.5, -15240)
  Copy logic: Copy rightward (positive X-axis) 16 times at 305 mm spacing.
- Lower longitudinal rebars:
  Initial rebar: from (105.5, -164.5, 0) to (105.5, -164.5, -15240)
  Copy logic: Copy rightward (positive X-axis) 16 times at 305 mm spacing.

Fig. B2. User Input in Case 01

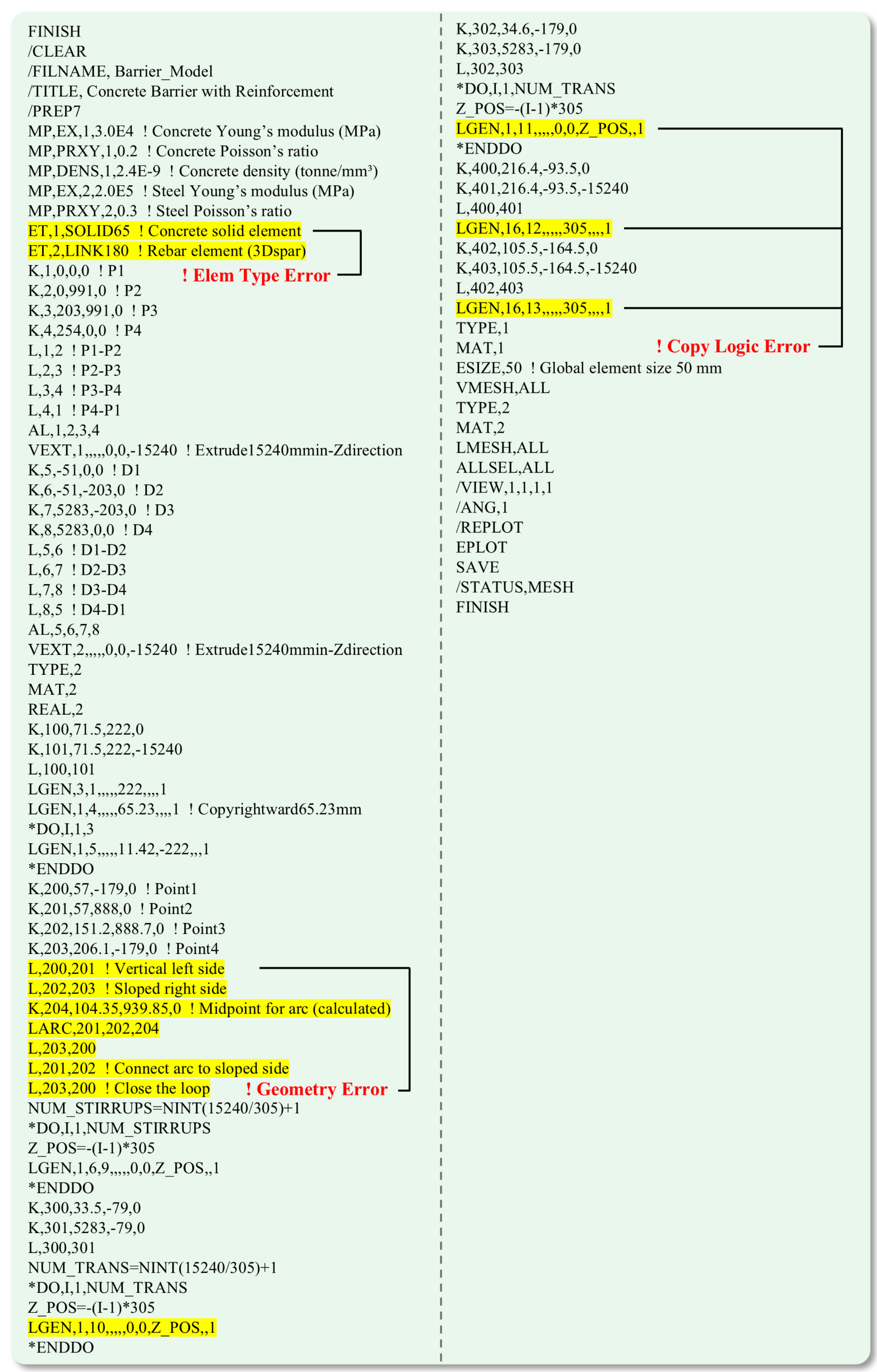

```
FINISH
/CLEAR
/FILNAME, Barrier_Model
/TITLE, Concrete Barrier with Reinforcement
/PREP7
MP,EX,1,3.0E4  ! Concrete Young's modulus (MPa)
MP,PRXY,1,0.2  ! Concrete Poisson's ratio
MP,DENS,1,2.4E-9  ! Concrete density (tonne/mm³)
MP,EX,2,2.0E5  ! Steel Young's modulus (MPa)
MP,PRXY,2,0.3  ! Steel Poisson's ratio
ET,1,SOLID65  ! Concrete solid element
ET,2,LINK180  ! Rebar element (3Dspar)
K,1,0,0,0  ! P1                ! Elem Type Error
K,2,0,991,0  ! P2
K,3,203,991,0  ! P3
K,4,254,0,0  ! P4
L,1,2  ! P1-P2
L,2,3  ! P2-P3
L,3,4  ! P3-P4
L,4,1  ! P4-P1
AL,1,2,3,4
VEXT,1,,,,,0,0,-15240  ! Extrude15240mmin-Zdirection
K,5,-51,0,0  ! D1
K,6,-51,-203,0  ! D2
K,7,5283,-203,0  ! D3
K,8,5283,0,0  ! D4
L,5,6  ! D1-D2
L,6,7  ! D2-D3
L,7,8  ! D3-D4
L,8,5  ! D4-D1
AL,5,6,7,8
VEXT,2,,,,,0,0,-15240  ! Extrude15240mmin-Zdirection
TYPE,2
MAT,2
REAL,2
K,100,71.5,222,0
K,101,71.5,222,-15240
L,100,101
LGEN,3,1,,,,,222,,,,1
LGEN,1,4,,,,,65.23,,,,1  ! Copyrightward65.23mm
*DO,I,1,3
LGEN,1,5,,,,,11.42,-222,,,1
*ENDDO
K,200,57,-179,0  ! Point1
K,201,57,888,0  ! Point2
K,202,151.2,888.7,0  ! Point3
K,203,206.1,-179,0  ! Point4
L,200,201  ! Vertical left side
L,202,203  ! Sloped right side
K,204,104.35,939.85,0  ! Midpoint for arc (calculated)
LARC,201,202,204
L,203,200
L,201,202  ! Connect arc to sloped side
L,203,200  ! Close the loop       ! Geometry Error
NUM_STIRRUPS=NINT(15240/305)+1
*DO,I,1,NUM_STIRRUPS
Z_POS=-(I-1)*305
LGEN,1,6,9,,,,,0,0,Z_POS,,1
*ENDDO
K,300,33.5,-79,0
K,301,5283,-79,0
L,300,301
NUM_TRANS=NINT(15240/305)+1
*DO,I,1,NUM_TRANS
Z_POS=-(I-1)*305
LGEN,1,10,,,,,0,0,Z_POS,,1
*ENDDO
K,302,34.6,-179,0
K,303,5283,-179,0
L,302,303
*DO,I,1,NUM_TRANS
Z_POS=-(I-1)*305
LGEN,1,11,,,,,0,0,Z_POS,,1
*ENDDO
K,400,216.4,-93.5,0
K,401,216.4,-93.5,-15240
L,400,401
LGEN,16,12,,,,,305,,,,1
K,402,105.5,-164.5,0
K,403,105.5,-164.5,-15240
L,402,403
LGEN,16,13,,,,,305,,,,1
TYPE,1
MAT,1                          ! Copy Logic Error
ESIZE,50  ! Global element size 50 mm
VMESH,ALL
TYPE,2
MAT,2
LMESH,ALL
ALLSEL,ALL
/VIEW,1,1,1,1
/ANG,1
/REPLOT
EPLOT
SAVE
/STATUS,MESH
FINISH
```

Fig. B3. End-to-End APDL Generation by DeepSeek-V3.2 in Case 01

## User Prompt: Barrier & Deck Modeling Specifications

Task: Generate complete APDL code for barrier modeling based on the user description.
Output: Generate the COMPLETE APDL code for the barrier model described above.
Requirements:

- Code must be complete, executable, and free of placeholders.
- Include all necessary engineering modeling steps.
- Apply strict, proper APDL native commands and syntax.
- Include descriptive comments explaining key operations.
- Ensure the entire script is organized, indented, and readable.

**I. Basic Modeling Parameters**

Coordinate origin (0, 0, 0): Located at the intersection of the barrier back face and the top surface of the bridge deck.
Unit: millimeters (mm).
Total length (L): 15240 mm (model extends in the negative Z-axis direction).

**II. Solid Modeling**

1. Barrier Cross-Section Profile

Connect the following vertices in sequence to form a closed polyline, then extrude it 15240 mm along the Z-axis:

- P1: (0, 0, 0)
- P2: (0, 1118, 0)
- P3: (254, 1118, 0)
- P4: (254, 0, 0)

2. Deck Cross-Section Profile

Connect the following vertices in sequence to form a closed polyline, then extrude it 15240 mm along the Z-axis:

- D1: (-51, 0, 0)
- D2: (-51, -254, 0)
- D3: (4216, -254, 0)
- D4: (4216, 0, 0)

**III. Barrier Reinforcement**

1. Longitudinal Rebars

All longitudinal rebars have a length of 15240 mm (extending from Z=0 to Z=−15240).

- Left longitudinal rebar array:
  Initial rebar: from (67, 105, 0) to (67, 105, -15240)
  Copy logic: Copy upward (positive Y-axis) 3 times at 305 mm spacing.
- Right longitudinal rebar array:
  Positioning: Take the last rebar generated above, copy it rightward (positive X-axis) once at 120 mm spacing to obtain the starting rebar.
  Copy logic: Copy this starting rebar downward (negative Y-axis) 3 times at 305 mm spacing.

2. Barrier Stirrups

Single stirrup path points (closed or U-shape):

- (51, -200.5, 0)
- (51, 1036, 0)
- (203, 1036, 0)
- (203, -200.5, 0)

Distribution: Array along the negative Z-axis at 152 mm spacing to cover the full length.

**IV. Deck Reinforcement**

1. Transverse Rebars

- Upper transverse rebars: Straight line connecting (0, -74, 0) and (4216, -74, 0).
- Lower transverse rebars: Straight line connecting (0, -195, 0) and (4216, -195, 0).

Distribution: Both layers arrayed along the negative Z-axis at 305 mm spacing.

2. Longitudinal Rebars

All longitudinal rebars have a length of 15240 mm (extending from Z=0 to Z=−15240).

- Upper longitudinal rebars:
  Initial rebar: from (34.8, -92, 0) to (34.8, -92, -15240)
  Copy logic 1: Copy rightward 5 times at 203 mm spacing.
  Copy logic 2: Based on the latest rebar generated, continue copying rightward 7 times at 305 mm spacing.
- Lower longitudinal rebars:
  Initial rebar: from (34.8, -179, 0) to (34.8, -179, -15240)
  Copy logic: Copy rightward (positive X-axis) 13 times at 305 mm spacing.

Fig. B4. User Input in Case 02

```
FINISH
/CLEAR
/FILNAME, Barrier_Model
/TITLE, Bridge Barrier with Reinforcement
/PREP7
K,1,0,0,0  ! P1: Bottom back point
K,2,0,1118,0  ! P2: Top back point
K,3,254,1118,0  ! P3: Top front point
K,4,254,0,0  ! P4: Bottom front point
L,1,2   ! Back vertical line
L,2,3   ! Top horizontal line
L,3,4   ! Front vertical line
L,4,1   ! Bottom horizontal line
AL,1,2,3,4
VEXT,1,,,,,0,0,-L_total
K,5,-51,0,0  ! D1
K,6,-51,-254,0  ! D2
K,7,4216,-254,0  ! D3
K,8,4216,0,0  ! D4
L,5,6
L,6,7
L,7,8
L,8,5
AL,5,6,7,8
VEXT,2,,,,,0,0,-L_total
ET,1,LINK180  ──────  ! Elem Type Error
R,1,201.06  ! Real constant for rebar area (assuming 16mm diameter: π*(8^2)=201.06 mm²)
K,100,67,105,0
K,101,67,105,-L_total
L,100,101
LGEN,3,1,,,,,0,305,0,,1  ─┐  ! Param Error
LGEN,1,4,,,,,120,0,0,,1   │
LGEN,3,5,,,,,0,-305,0,,1 ─┘
K,200,51,-200.5,0
K,201,51,1036,0
K,202,203,1036,0
K,203,203,-200.5,0
L,200,201
L,201,202
L,202,203
n_stirrups = NINT(L_total/152) + 1
*DO,i,1,n_stirrups
z_offset = -(i-1)*152
LGEN,1,6,8,,,0,0,z_offset,,1  ──  ! Param Error
*ENDDO
K,300,0,-74,0
K,301,4216,-74,0
L,300,301
K,302,0,-195,0
K,303,4216,-195,0
L,302,303
n_transverse = NINT(L_total/305) + 1
*DO,i,1,n_transverse
z_offset = -(i-1)*305
LGEN,1,9,,,0,0,z_offset,,1  ! Copy upper transverse
LGEN,1,10,,,0,0,z_offset,,1  ! Copy lower transverse
*ENDDO
K,400,34.8,-92,0
K,401,34.8,-92,-L_total
L,400,401
LGEN,5,11,,,,,203,0,0,,1  ─┐  ! Undefined Line ID
LGEN,7,15,,,,,305,0,0,,1  ─┘
K,402,34.8,-179,0
K,403,34.8,-179,-L_total
L,402,403
LGEN,13,22,,,,,305,0,0,,1
LSEL,ALL
LATT,1,1,1
LESIZE,ALL,,,1! One element per line
LMESH,ALL
ALLSEL
VATT,2,2,2! Assuming material 2 for concrete
ESIZE,50  ! Element size 50 mm
VMESH,ALL
ALLSEL,ALL
EPLOT
SAVE
FINISH
```

Fig. B5. End-to-End APDL Generation by DeepSeek-V3.2 in Case 02

## User Prompt: Barrier & Deck Modeling Specifications

Task: Generate complete APDL code for barrier modeling based on the user description.
Output: Generate the COMPLETE APDL code for the barrier model described above.
Requirements:

- ➢ Code must be complete, executable, and free of placeholders.
- ➢ Include all necessary engineering modeling steps.
- ➢ Apply strict, proper APDL native commands and syntax.
- ➢ Include descriptive comments explaining key operations.
- ➢ Ensure the entire script is organized, indented, and readable.

**I. Basic Modeling Parameters**

Coordinate origin (0, 0, 0): Located at the intersection of the barrier back face and the top surface of the bridge deck.
Unit: millimeters (mm).
Total length (L): 15240 mm (model extends in the negative Z-axis direction).

**II. Solid Modeling**

1. Barrier Cross-Section Profile

Connect the following 6 vertices in sequence to form a closed polyline, then extrude it 15240 mm along the Z-axis:

- P1: (0,0,0) — Bottom back point
- P2: (0,813,0) — Top back point
- P3: (152,813,0) — Top edge point
- P4: (203,330,0) — Slope turning point 1
- P5: (381,76,0) — Slope turning point 2 (barrier footing point)
- P6: (381,0,0) — Bottom front point

2. Deck Cross-Section Profile

Connect the following vertices in sequence to form a closed polyline, then extrude it 15240 mm along the Z-axis:

- D1: (−38,0,0)
- D2: (−38,−254,0)
- D3: (4407,−254,0)
- D4: (4407,0,0)

**III. Barrier Reinforcement**

1. Longitudinal Rebars

All longitudinal rebars have a length of 15240 mm (extending from Z=0 to Z=−15240).

- ○ Left longitudinal rebar array:
  Initial rebar: Start point (59.4,210,0)
  Copy logic: Copy upward (positive Y-axis) 3 times at 178 mm spacing.
- ○ Right longitudinal rebar array:
  Positioning: Take the last rebar generated above, copy it rightward (positive X-axis) once at 39.2 mm spacing to obtain the starting rebar.
  Copy logic: Copy this starting rebar downward and rightward 3 times; displacement increment: 18.8 mm horizontally right, 178 mm vertically downward (to keep the rebar mesh parallel to the complex sloped face).

2. Barrier Stirrups

This part consists of three components, arranged at intervals of 203 mm along the length :

- Stirrup 1 (Vertical rebar): Straight line from (46,−200,0) to (46,765.7,0).
- Stirrup 2 (Composite rebar):Path: (46,766,0)→(46,140.9,0), then connect to (176.3,145.2,0), and finally to (110.7,766,0).Special requirement: Create a downward concave arc between (46,140.9) and (176.3,145.2).
- Stirrup 3 (Tie rebar): Path: (46,473.9,0)→(332.8,62.5,0)→(335,−200,0). Distribution: Array along the negative Z-axis at 305 mm spacing to cover the full length.

**IV. Deck Reinforcement**

1. Transverse Rebars

- ○ Standard transverse rebars: Connect (21.5,−59,0) and (4407,−59,0), spaced at 305 mm along the Z-axis.
- ○ Reinforced dense rebars: Path: Connect (21.5,−59,−152.5) and (1359,−59,−152.5). Distributed at 305 mm spacing along the Z-axis.
- ○ Lower transverse rebars: Connect (21.5,−208,0) and (4407,−208,0), spaced at 305 mm along the Z-axis.

2. Longitudinal Rebars

All longitudinal rebars have a length of 15240 mm.

- ○ Upper longitudinal rebars: Initial rebar: Start (21.5,−75,0), end (21.5,−75,−15240).
  Copy logic: First copy rightward 5 times at 356 mm spacing; then based on the newly generated rebars, copy rightward another 5 times at 508 mm spacing.
- ○ Lower longitudinal rebars: Initial rebar: Start (21.5,−192,0), end (21.5,−192,−15240).Copy logic: Copy rightward 8 times at 508 mm spacing.

Fig. B6. User Input in Case 03

```
FINISH
/CLEAR
/FILNAME, Barrier_Model
/TITLE, Bridge Barrier with Reinforcement
/PREP7
MP,EX,1,3.0e4  ! Concrete Young's modulus
MP,PRXY,1,0.2  ! Concrete Poisson's ratio
MP,DENS,1,2.4e-9  ! Concrete density
MP,EX,2,2.0e5  ! Steel Young's modulus
MP,PRXY,2,0.3  ! Steel Poisson's ratio
ET,1,SOLID186   ! 3D solid element for concrete
ET,2,LINK180  ! 3D spar element for rebar
K,1,0,0,0  ! P1
K,2,0,813,0  ! P2                    ! Elem Type Error
K,3,152,813,0  ! P3
K,4,203,330,0  ! P4
K,5,381,76,0  ! P5
K,6,381,0,0  ! P6
L,1,2  ! P1-P2
L,2,3  ! P2-P3
L,3,4  ! P3-P4
L,4,5  ! P4-P5
L,5,6  ! P5-P6
L,6,1  ! P6-P1 (closes the profile)
AL,ALL
VEXT,ALL,,,,,0,0,-15240  ! Extrude 15240 mm in -Z direction
K,100,-38,0,0  ! D1
K,101,-38,-254,0  ! D2
K,102,4407,-254,0  ! D3
K,103,4407,0,0  ! D4
L,100,101  ! D1-D2
L,101,102  ! D2-D3
L,102,103  ! D3-D4
L,103,100  ! D4-D1
AL,ALL
VEXT,ALL,,,,,0,0,-15240
SECTYPE,1,LINK
SECDATA,201.06
K,200,59.4,210,0
K,201,59.4,210,-15240
L,200,201 ! Create first rebar
LGEN,3,ALL,,,,,0,178,0,,1
LSEL,S,LINE,,4  ! Select the last rebar
LGEN,1,ALL,,,,,39.2,0,0,,1
LSEL,S,LINE,,5                  ! Param Error
*DO,I,1,3
LGEN,1,ALL,,,,,18.8,-178,0,,1
*ENDDO
SECTYPE,2,LINK
SECDATA,113.1
K,300,46,-200,0
K,301,46,765.7,0
L,300,301
K,302,46,766,0
K,303,46,140.9,0
K,304,176.3,145.2,0
K,305,110.7,766,0
L,302,303  ! First vertical segment
L,304,305  ! Top diagonal segment
CIRCLE,111.15,143.05,0,65.15,0,360
L,303,106  ! Connect to arc start
L,107,304   ! Connect from arc end
K,306,46,473.9,0                 ! Geometry Error
K,307,332.8,62.5,0
K,308,335,-200,0
L,306,307  ! First segment
L,307,308  ! Second segment
ALLSEL
LSEL,S,LINE,,6,14
```

```
NSECT = 15240/203  ! Calculate number of sections
NSECT = NSECT – 1  ! Subtract initial position
*DO,I,1,NSECT
ZPOS = -203*I  ! Negative Z direction
LGEN,1,ALL,,,,,0,0,ZPOS,,1
*ENDDO
SECTYPE,3,LINK
SECDATA,314.16
K,400,21.5,-59,0
K,401,4407,-59,0
L,400,401 ! Create first transverse rebar
LSEL,S,LINE,,LAST
NZSEC = 15240/305
NZSEC = NZSEC - 1
*DO,I,1,NZSEC
ZPOS = -305*I  ! Negative Z direction
LGEN,1,ALL,,,,,0,0,ZPOS,,1
*ENDDO
K,402,21.5,-59,-152.5
K,403,1359,-59,-152.5
L,402,403
LSEL,S,LINE,,LAST
*DO,I,1,NZSEC
ZPOS = -305*I  ! Negative Z direction
LGEN,1,ALL,,,,,0,0,ZPOS,,1
*ENDDO
K,404,21.5,-208,0
K,405,4407,-208,0
L,404,405
LSEL,S,LINE,,LAST
*DO,I,1,NZSEC
ZPOS = -305*I  ! Negative Z direction
LGEN,1,ALL,,,,,0,0,ZPOS,,1
*ENDDO                        ! Logic error
K,500,21.5,-75,0
K,501,21.5,-75,-15240
L,500,501  ! Create first upper longitudinal rebar
LSEL,S,LINE,,LAST
LGEN,5,ALL,,,,,356,0,0,,1
LSEL,S,LINE,,2,6
LGEN,5,ALL,,,,,508,0,0,,1
K,503,21.5,-192,-15240
K,502,21.5,-192,0
L,502,503
LSEL,S,LINE,,LAST
LGEN,8,ALL,,,,,508,0,0,,1
ALLSEL
VSEL,S,VOLU,,1,2
VATT,1,1,1
LSEL,S,LINE,,1,5
LATT,2,1,2,,,
LSEL,S,LINE,,
LSEL,A,LINE,,15,1000 LATT,2,2,2,,,
LSEL,S,LINE,,100,500 ! Select deck transverse rebars (approximate range)
LATT,2,3,2,,,                 ! Param Error
Type2,Section3
LSEL,S,LINE,,600,700
LATT,2,3,2,,,
ALLSEL
ESIZE,100
MSHAPE,1,3D
MSHKEY,0
VMESH,ALL
LSEL,S,LINE,,1,1000
ESIZE,50
LMESH,ALL
ALLSEL
```

```
NSEL,S,LOC,Y,-254
D,ALL,ALL,0
NSEL,S,LOC,Z,0
DSYM,SYMM,Z
NSEL,S,LOC,Z,-15240
DSYM,SYMM,Z
ALLSEL
ACEL,0,9800,0
/SOLU
ANTYPE,0
NLGEOM,ON
NSUBST,10
OUTRES,ALL,ALL
SOLVE
/POST1
PLNSOL,U,SUM
PLNSOL,S,EQV
SAVE,Barrier_Model,db
FINISH
```

Fig. B7. End-to-End APDL Generation by DeepSeek-V3.2 in Case 03

## User Prompt: Barrier & Deck Modeling Specifications

Task: Generate complete APDL code for barrier modeling based on the user description.
Output: Generate the COMPLETE APDL code for the barrier model described above.
Requirements:
- Code must be complete, executable, and free of placeholders.
- Include all necessary engineering modeling steps.
- Apply strict, proper APDL native commands and syntax.
- Include descriptive comments explaining key operations.
- Ensure the entire script is organized, indented, and readable.

**I. Basic Modeling Parameters**
Coordinate system origin (0, 0, 0): Located at the intersection of the barrier's back face and the top surface of the bridge deck.
Unit: millimeters (mm).
Total length (L): 15240 mm (model extends in the negative Z-axis direction).

**II. Solid Modeling**
1. Barrier Cross-Section Profile
Connect the following vertices in sequence to form a closed rectangular polyline, then extrude it by 15240 mm along the Z-axis:
- P1: (0,0,0) — Bottom back point
- P2: (0,812,0) — Top back point
- P3: (254,812,0) — Top front point
- P4: (254,0,0) — Bottom front point

2. Deck Cross-Section Profile
Connect the following vertices to form a closed cross-section (calculated based on coordinates, with a thickness of 204 mm), then extrude it by 15240 mm:
- D1: (0,−204,0)
- D2: (4826,−204,0)
- D3: (4826,0,0)

**III. Barrier Reinforcement**
1. Longitudinal Rebars
All longitudinal rebars have a length of 15240 mm (extending from Z=0 to Z=−15240).
- Left Left longitudinal rebar array:
  Initial rebar coordinates: Start point (75,51,0)
  Copy logic: Copy upward (positive Y-axis direction) 3 times with a spacing of 203 mm.
- Right longitudinal rebar array:
  Positioning: Select the last rebar generated above, copy it rightward (positive X-axis direction) once with a spacing of 104 mm to obtain the starting rebar.
  Copy logic: Select this starting rebar, copy it downward (negative Y-axis direction) 3 times with a spacing of 203 mm.

2. Barrier Stirrups
- Single stirrup path points: (59,−158,0)→(59,667.9,0)→(195,667.9,0)→(195,−158,0)→(380,−158,0)Distribution method: Array arrangement along the negative Z-axis direction with a spacing of 178 mm.

**IV. Deck Reinforcement**
1. Transverse Rebars
- Standard transverse rebars: Connect (59,−46,0) and (4826,−46,0), with a spacing of 229 mm along the Z-axis.
- Reinforced dense rebars: Path: Connect (59,−46,−114.5) and (1778,−46,−114.5). Distribution: Array arrangement along the Z-axis with a spacing of 229 mm.
- Lower transverse rebars: Connect (59,−158,0) and (4826,−158,0), with a spacing of 229 mm along the Z-axis.

2. Longitudinal Rebars
All longitudinal rebars have a length of 15240 mm.
- Upper longitudinal rebars:
  Initial rebar start point: (75,−62,0), end point (75,−62,−15240).
  Copy logic: Copy rightward (positive X-axis direction) 15 times with a spacing of 305 mm.
- Lower longitudinal rebars:
  Initial rebar start point: (75,−142,0), end point (75,−142,−15240).
  Copy logic: Copy rightward (positive X-axis direction) 15 times with a spacing of 305 mm.

Fig. B8. User Input in Case 04

```
FINISH
/CLEAR
/FILNAME, Barrier_Model
/TITLE, Bridge Barrier with Reinforcement
/PREP7
ET,1,SOLID185  ! For concrete solid elements
ET,2,LINK180  ! For rebar elements            ! Elem Type Error
MP,EX,1,30E3  ! Young's modulus (MPa)
MP,PRXY,1,0.2  ! Poisson's ratio
MP,DENS,1,2.4E-9  ! Density (tonne/mm^3)
MP,EX,2,200E  ! Young's modulus (MPa)
MP,PRXY,2,0.3  ! Poisson's ratio
MP,DENS,2,7.85E-9  ! Density (tonne/mm^3)
K,1,0,0,0  ! P1: Bottom back point
K,2,0,812,0  ! P2: Top back point
K,3,254,812,0  ! P3: Top front point
K,4,254,0,0  ! P4: Bottom front point
L,1,2  ! Line from P1 to P2
L,2,3  ! Line from P2 to P3
L,3,4  ! Line from P3 to P4
L,4,1  ! Line from P4 to P1
AL,1,2,3,4  ! Create area from lines 1-4
VEXT,1,,,,,0,0,-15240  ! Extrude area 1 along Z-axis
K,5,0,0,0  ! D1: Starting point (same as P1)
K,6,0,-204,0  ! D2          ! Duplicate Point
K,7,4826,-204,0  ! D3
K,8,4826,0,0  ! D4
L,5,6  ! Line D1-D2
L,6,7  ! Line D2-D3
L,7,8  ! Line D3-D4
L,8,5  ! Line D4-D1
AL,5,6,7,8  ! Create area from deck lines
VEXT,2,,,,,0,0,-15240  ! Extrude deck area along Z-axis
SECTYPE,1,LINK  ! Define rebar section type
SECDATA,16  ! Rebar diameter 16 mm (adjust as needed)
K,100,75,51,0  ! Start point
K,101,75,51,-15240  ! End point          ! Param Error
L,100,101  ! Create rebar line
LGEN,3,LAST,,,,203,,,1 ! Copy upward 3 times
LSEL,S,LOC,X,75
LSEL,R,LOC,Y,51+203*2  ! Select top rebar of left array
LGEN,1,ALL,,,,104,,,1  ! Copy rightward 104 mm
ALLSEL,ALL
LSEL,S,LOC,X,75+104
LSEL,R,LOC,Y,51+203*2
LGEN,3,ALL,,,,-203,,,1 ! Copy downward 3
ALLSEL,ALL
K,200,59,-158,0  ! Point 1
K,201,59,667.9,0  ! Point 2
K,202,195,667.9,0  ! Point 3
K,203,195,-158,0  ! Point 4
K,204,380,-158,0  ! Point 5
L,200,201  ! Line 1-2
L,201,202  ! Line 2-3
L,202,203  ! Line 3-4
L,203,204  ! Line 4-5
CM,STIRRUP1,LINE  ! Create component for first stirrup
LGEN,86,STIRRUP1,,,,,,-178,0  ! Copy 86 times with 178
mm spacing
K,300,59,-46,0  ! Start point
K,301,4826,-46,0  ! End point
L,300,301  ! Create rebar line
LGEN,67,LAST,,,,,,-229,0  ! Copy 67 times with 229 mm
spacing                                  ! Param Error
K,302,59,-46,-114.5  ! Start point
K,303,1778,-46,-114.5  ! End point
302,303  ! Create rebar line
L, LGEN,67,LAST,,,,,,-229,0  ! Copy 67 times with 229
mm spacing
K,304,59,-158,0   ! Start point
K,305,4826,-158,0  ! End point
L,304,305  ! Create rebar line
LGEN,67,LAST,,,,,,-229,0  ! Copy 67 times with 229 mm
spacing
K,400,75,-62,0  ! Start point
K,401,75,-62,-15240  ! End point
L,400,401  ! Create rebar line
LGEN,15,LAST,,,,305,,,1  ! Copy rightward 15 times with
305 mm spacing
K,402,75,-142,0  ! Start point
K,403,75,-142,-15240  ! End point
L,402,403  ! Create rebar line
LGEN,15,LAST,,,,305,,,1  ! Copy rightward 15 times with
305 mm spacing
SAVE,Barrier_Model,db
FINISH                                   ! Param Error
/POST1
/VIEW,1,1,1,1
/ANG,1
/REPLOT
EPLOT
VPLOT
FINISH
```

Fig. B9. End-to-End APDL Generation by DeepSeek- V3.2 in Case 04

## User Prompt: Barrier & Deck Modeling Specifications

Task: Generate complete APDL code for barrier modeling based on the user description.
Output: Generate the COMPLETE APDL code for the barrier model described above.
Requirements:

- Code must be complete, executable, and free of placeholders.
- Include all necessary engineering modeling steps.
- Apply strict, proper APDL native commands and syntax.
- Include descriptive comments explaining key operations.
- Ensure the entire script is organized, indented, and readable.

**I. Basic Modeling Parameters**
Coordinate origin (0, 0, 0): Intersection between the vertical back face of the barrier and the top surface of the bridge deck.
Unit: millimeters (mm).
Total length (L): 15240 mm (model extends in the negative Z-axis direction).

**II. Solid Modeling**
<u>1. Barrier Cross-Section Profile</u>
Connect the following vertices in sequence to form a closed polyline, then extrude along the Z-axis to Z=−15240:

- P1: (0,0,0) — Bottom back point
- P2: (0,914,0) — Top back point
- P3: (152,914,0) — Top edge point
- P4: (330,0,0) — Bottom front point (slope junction))

<u>2. Deck Cross-Section Profile</u>
Connect the following vertices in sequence to form a closed polyline, then extrude along the Z-axis to Z=−15240:

- D1: (−38,0,0)
- D2: (−38,−203,0)
- D3: (4072,−203,0)
- D4: (4072,0,0)

**III. Barrier Reinforcement**
<u>1. Longitudinal Rebars</u>
All longitudinal rebars have a length of 15240 mm.

- Left (vertical face) rebar array:
  First rebar: start (59.8,19,0), end (59.8,19,−15240)
  Copy logic: Copy upward (positive Y-axis) 4 times at 203 mm spacing.
- Right (sloped face) rebar array:
  Starting rebar: Take the top rebar from the left array, copy rightward (positive X-axis) once at 47.6 mm spacing.
  First rebar: start (107.4,831,0), end (107.4,831,−15240)
  Copy logic: Copy downward and rightward 4 times; displacement increment: 39.5 mm horizontally right, 203 mm vertically downward..

<u>2. Barrier Stirrups</u>
One set every 152 mm, consisting of two components:

- Stirrup 1 (Upper):Path: (50,0,0)→(50,836.1,0)→(116.2,836.1,0)→(277.8,0,0)Special treatment: Create an upward convex arc between (50,836.1) and (116.2,836.1).
- Stirrup 2 (Lower anchorage):Path: (72.6,−170.8,0)→(72.6,222,0)→(181.6,222,0)→(181.6,−170.8,0)→(305,−170.8,0)Special treatment: Create an upward convex arc between (72.6,222) and (181.6,222).

**IV. Deck Reinforcement**
<u>1. Transverse Rebars</u>

- Standard upper transverse rebars: start (0,−59,0), end (4072,−59,0); spaced at 305 mm along Z-axis.
- Dense transverse rebars: start (0,−59,−152.5), end (1024,−59,−152.5); spaced at 305 mm along Z-axis.
- Lower transverse rebars: start (0,−163,0), end (4072,−163,0); spaced at 305 mm along Z-axis.

<u>2. Longitudinal Rebars</u>
All longitudinal rebars have a length of 15240 mm (extending from Z=0 to Z=−15240).

- Upper Longitudinal Rebars
  First rebar: start (0,−73.5,0), end (0,−73.5,−15240)
  Copy logic 1: Copy rightward 6 times at 229 mm spacing.
  Copy logic 2: Based on the latest rebar, continue copying rightward 7 times at 305 mm spacing.
- Lower Longitudinal Rebars
  First rebar: start (0,−147,0), end (0,−147,−15240)
  Copy logic: Copy rightward 12 times at 305 mm spacing.

Fig. B10. User Input in Case 05

```
FINISH
/CLEAR
/FILNAME, Barrier_Model
/TITLE, Bridge Barrier with
Reinforcement
/PREP7
ET,1,SOLID185  ! Elem Type Error
ET,2,LINK180
MP,EX,1,3.0E4MP,PRXY,1,0.2MP,
DENS,1,2.4E-9
MP,EX,2,2.0E5
MP,PRXY,2,0.3
MP,DENS,2,7.85E-9
R,1,201.06
R,2,113.10
K,1,0,0,0  ! P1
K,2,0,914,0  ! P2
K,3,152,914,0 ! P3
K,4,330,0,0  ! P4
L,1,2  ! Vertical back face
L,2,3  ! Top horizontal
L,3,4  ! Sloped front face
L,4,1  ! Bottom horizontal
AL,1,2,3,4
VEXT,1,,,,,0,0,-15240
K,5,-38,0,0  ! D1
K,6,-38,-203,0  ! D2
K,7,4072,-203,0  ! D3
K,8,4072,0,0  ! D4
L,5,6
L,6,7
L,7,8
L,8,5
AL,5,6,7,8
VEXT,2,,,,,0,0,-15240
TYPE,2
REAL,1
MAT,2
K,100,59.8,19,0
K,101,59.8,19,-15240
L,100,101
LESIZE,ALL,,,10
LMESH,ALL
*DO,I,1,4
LGEN,2,ALL,,,,,0,203,0,,,,1
*ENDDO
LSEL,S,LOC,Y,831
LGEN,2,ALL,,,,,47.6,0,0,,,,1
ALLSEL
K,200,107.4,831,0
K,201,107.4,831,-15240
L,200,201
LESIZE,ALL,,,10
LMESH,ALL
*DO,I,1,4
LGEN,2,ALL,,,,,39.5,-203,0,,,,1
*ENDDO        ! Param Error
REAL,2
K,300,50,0,0
K,301,50,836.1,0
K,302,116.2,836.1,0
K,303,277.8,0,0
L,300,301
L,302,303
K,304,83.1,881.1,0
LARC,301,302,304
L,303,300  ! Geometry Error
LESIZE,ALL,,,4
LMESH,ALL
K,400,72.6,-170.8,0
K,401,72.6,222,0
K,402,181.6,222,0
K,403,181.6,-170.8,0
K,404,305,-170.8,0
L,400,401
L,403,404
L,404,400
K,405,127.1,272,0
LARC,401,402,405
L,402,403
LESIZE,ALL,,,4
LMESH,ALL
ALLSEL
LSEL,S,TYPE,,2 ! Select all
NSEL,S,LOC,Z,0
NSEL,R,LOC,Y,0,1000
NUM_COPIES=15240/152
*DO,I,1,NUM_COPIES
EGEN,2,ALL,ALL,,,,,0,0,-152*I
*ENDDO
REAL,1
K,500,0,-59,0
K,501,4072,-59,0
L,500,501
LESIZE,ALL,,,20
LMESH,ALL
NUM_T=15240/305
*DO,I,1,NUM_T
LGEN,2,ALL,,,,,0,0,-305*I,,,,1
*ENDDO
K,502,0,-59,-152.5
K,503,1024,-59,-152.5
L,502,503
LESIZE,ALL,,,10
LMESH,ALL
*DO,I,1,NUM_T
LGEN,2,ALL,,,,,0,0,-305*I,,,,1
*ENDDO
K,504,0,-163,0
K,505,4072,-163,0
L,504,505
LESIZE,ALL,,,20
LMESH,ALL
*DO,I,1,NUM_T
LGEN,2,ALL,,,,,0,0,-305*I,,,,1
*ENDDO
K,600,0,-73.5,0
K,601,0,-73.5,-15240
L,600,601
LESIZE,ALL,,,10
LMESH,ALL
*DO,I,1,6
LGEN,2,ALL,,,,,229*I,0,0,,,,1
*ENDDO
LSEL,S,LOC,X,229*6
*DO,I,1,7
LGEN,2,ALL,,,,,229*6+305*I,0,0,,,,1
*ENDDO
ALLSEL
K,700,0,-147,0
K,701,0,-147,-15240
L,700,701
LESIZE,ALL,,,10
LMESH,ALL
*DO,I,1,12
LGEN,2,ALL,,,,,305*I,0,0,,,,1
*ENDDO        ! Logic error
ALLSEL
TYPE,1
MAT,1
REAL,3
VSEL,S,VOLU,,1
ESIZE,50  ! Element size 50 mm
VMESH,ALL
VSEL,S,VOLU,,2
ESIZE,100 ! Element size 100 mm
VMESH,ALL
ALLSEL
SAVE
FINISH
/POST1
/VIEW,1,1,1,1
/ANG,1
/REPLOT
/ESHAPE,1
EPLOT
FINISH
```

Fig. B11. End-to-End APDL Generation by DeepSeek-V3.2 in Case 05